\documentclass{article}

\usepackage{arxiv}

\usepackage[utf8]{inputenc} 
\usepackage[T1]{fontenc}    
\usepackage{booktabs}       
\usepackage{amsfonts}       
\usepackage{nicefrac}       
\usepackage{amsmath} 
\usepackage{amsthm}
\usepackage{graphicx}
\usepackage{natbib}
\usepackage{doi}
\usepackage{dsfont}
\usepackage{mathtools} 
\usepackage{nicematrix}
\usepackage{extarrows} 
\usepackage{hyperref}

\title{Regularized Linear Regression for Binary Classification}

\date{}

\newif\ifuniqueAffiliation
\uniqueAffiliationtrue

\ifuniqueAffiliation 
\author{ Danil Akhtiamov$^*$ \\
	Department of Computing and Mathematical Sciences\\
	California Institute of Technology\\
	Pasadena, CA 91125 \\
	\texttt{dakhtiam@caltech.edu} \\
	\And
	Reza Ghane$^*$ \\
	Department of Electrical Engineering\\
	California Institute of Technology\\
	Pasadena, CA 91125 \\
	\texttt{rghanekh@caltech.edu} \\
	\AND
        Babak Hassibi \\
	Department of Electrical Engineering\\
	California Institute of Technology\\
	Pasadena, CA 91125 \\
	\texttt{hassibi@caltech.edu} \\
}
\else
\usepackage{authblk}
\fi

\renewcommand{\shorttitle}{REGULARIZED LINEAR REGRESSION FOR BINARY
CLASSIFICATION}

\newtheorem{theorem}{Theorem}
\newtheorem{remark}{Remark}
\newtheorem{lemma}{Lemma}

\newtheorem{definition}{Definition}

\begin{document}

\maketitle

\def\thefootnote{*}\footnotetext{Equal contribution}

\begin{abstract}

Regularized linear regression is a promising approach for binary classification problems in which the training set has noisy labels since the regularization term can help to avoid interpolating the mislabeled data points. In this paper we provide a systematic study of the effects of the regularization strength on the performance of linear classifiers that are trained to solve binary classification problems by minimizing a regularized least-squares objective. We consider the over-parametrized regime and assume that the classes are generated from a Gaussian Mixture Model (GMM) where a fraction $c<\frac{1}{2}$ of the training data is mislabeled. Under these assumptions, we rigorously analyze the classification errors resulting from the application of ridge, $\ell_1$, and $\ell_\infty$ regression. In particular, we demonstrate that ridge regression invariably improves the classification error. We prove that $\ell_1$ regularization induces sparsity and observe that in many cases one can sparsify the solution by up to two orders of magnitude without any considerable loss of performance, even though the GMM has no underlying sparsity structure. For $\ell_\infty$ regularization we show that, for large enough regularization strength, the optimal weights concentrate around two values of opposite sign. We observe that in many cases the corresponding "compression" of each weight to a single bit leads to very little loss in performance. These latter observations can have significant practical ramifications.

\end{abstract}

\section{INTRODUCTION}

As the usage of machine learning models becomes more prevalent, the need for efficiently storing these models and for guaranteeing their performance in the face of noisy training data becomes increasingly vital. With the advent of LLMs, which are comprised of many billions of weights, the necessity for reliable compression schemes becomes ever more critical. Thus, the question is raised, are there methods that allow us to compress the weights of a deep neural net without compromising a lot on the performance? In this paper, we would like to take baby steps towards addressing this question via the analysis of the toy problem of regularized linear regression for binary classification. Regularization can be used to avoid fitting corrupted data, thereby improving performance, as well as to favor solutions with desired properties, such as sparsity or compressibility. Through theory, intuitive arguments, and numerical simulations, we demonstrate that regularization can help the generalization performance on noise-corrupted data sets, as well as reduce the number of model parameters by orders of magnitude without significant loss in performance.

\section{RELATED WORKS AND OUR CONTRIBUTION}

There has been a recent surge of results in binary classification that provide a sharp analysis of a variety of methods tailored to different models (see, e.g., \cite{thrampoulidis2018precise,thrampoulidis2020theoretical, HH17,candessur2018,cansur19,ka20,salehi2019impact,tpt20,zat19,mrsy19,mklz20,lolas20} and the references therein). These works typically pose the over-parameterized binary classification problem as an optimization problem and employ either the Convex Gaussian Min-max Theorem (CGMT) \cite{thrampoulidis2015regularized,thrampoulidis2018precise,stojnic2013framework,gordon1985some, loureiro2021learning} or the Approximate Message Passing (AMP) (see, e.g, \cite{donoho2009message, bayati2011dynamics, javanmard2013state}) approach to obtain formulas for the generalization error that involve solutions to a system of non-linear equations (in a small number of variables) that often do not admit closed-form expressions. These results follow a long line of work that deals with obtaining {\em sharp} high-dimensional asymptotics for convex optimization-based estimators. Most of the papers referred to above use some form of linear, logistic, or max-margin optimization. Most relevant to the scope of the present work are those that study binary classification through the lens of {\em regularized} linear regression which we highlight below.

\cite{loureiro2021learning} explores both binary and multi-class classification using an arbitrary convex loss and quadratic regularizer. Since our focus is on the effects of regularization, such as sparsification and compression, we instead study quadratic loss and arbitrary regularizer. They demonstrate how to reduce the analysis of the generalization error to finding fixed points of a low-dimensional system of equations using AMP. We should remark that we currently do not know how to use the CGMT framework to analyze the multi-class setting, a topic that is worthy of investigation in its own right---see e.g., \cite{thrampoulidis2020theoretical} for an attempt in this direction.

\cite{pmlr-v119-mignacco20a} uses Gaussian comparison inequalities to analyze the binary classification error for arbitrary loss functions with $\ell_2$ regularization. The main differences between our works are:

$\bullet$ The loss considered in \cite{pmlr-v119-mignacco20a} is of the form $\mathcal{L}(y(a^T w +b))$, whereas ours is $\mathcal{L}(y-a^Tw)$, as we aim to study performance of the regression-based approaches to classification.

$\bullet$ We introduce corruption to the labels and analyze how it affects the generalization error. This seems more natural to do because one of the main reasons for explicit regularization is finding solutions that do not interpolate the data. 

$\bullet$ We consider arbitrary separable convex regularizers and, in particular, show that $\ell_1$ regularization allows one to find sparse classifiers and $\ell_\infty$ regularization can lead to $1$-bit compression of the solution. For simplicity, we have focused on quadratic loss so that we could highlight the effect of the regularizer.

$\bullet$ In the case of $\ell_2$ regularization we reduce the number of scalar parameters to find the generalization error from six in \cite{pmlr-v119-mignacco20a} to two. Furthermore, in the regime of strong regularization, i.e., $\lambda \to \infty$, we find a closed form expression for the generalization error and use it to conclude that an arbitrary corruption rate $c < 0.5$ can be annihilated with large enough regularization strength.

Other works have studied corruption in the labels.  Most notably \cite{chatterji2021finite} analyzes the performance of the max-margin algorithm for binary classification of linearly separable data in the presence of corruption in the labels. 
\section{PRELIMINARIES}
\subsection{Gaussian Mixture Model}\label{bin_class}

We consider a binary classification problem with two classes, where for class $i=1,2$ the feature vector $x\in\mathbb{R}^d$ is drawn at random from $\mathcal{N}(\mu_i, \Sigma_i)$, with $\mu_i \in\mathbb{R}^d$ the mean and $\Sigma_i\in\mathbb{R}^{d\times d}$ the covariance matrix, and where the label is chosen as $y=1$ for $i=1$ and $y=-1$ for $i=2$. How well a linear classifier performs depends on how "close" the mean vectors $\mu_1$ and $\mu_2$ are and what the structure of the covariance matrices is. In the sequel, we will largely assume that the covariances are isotropic ($\Sigma_1 = \Sigma_2 = \sigma^2I$). For the means, we will assume that their matching components are drawn iid from zero mean standard normal distributions with cross-correlation $-1\leq r\leq 1$. The cross-correlation $r$ allows one to control the distance between the cluster centers. For example, this means that for large $d$, $\|\mu_1\|^2 \approx \|\mu_2\|^2 \approx d$ and $\|\mu_1-\mu_2\|^2 \approx 2(1-r)d$. 

For simplicity of exposition, we treat only the case of equal class sizes in this paper, but it is straightforward to apply the same techniques to the case of unbalanced classes. What we mean by equal classes is that the number of training data points for each class is $\frac{n}{2}$ (for a total of $n$ training points) and that the probability of drawing an element from each class, in order to determine the generalization error, is $\frac{1}{2}$. In all our subsequent analysis we will assume that we are in the over-parametrized regime, i.e., $n<d$ (often $n\ll d)$. Finally, we will assume that a fraction $c<\frac{1}{2}$ of the training dataset is mislabeled. 

We will consider a linear classifier given by a weight vector $w\in\mathbb{R}^d$. In other words for a given feature vector $x$, we will declare that $x$ belongs to class 1 if $x^Tw>0$ and to class 2 if $x^Tw<0$. It is then straightforward to show the following result.
\begin{lemma}\label{lem:gen_error}
Given a weight vector $w$, and assuming the feature vectors are equally likely to be drawn from class 1 or class 2, the corresponding generalization error for the Gaussian mixture model with means $\mu_1$ and $\mu_2$ and covariance matrices $\Sigma_1, \Sigma_2$ is given by
\begin{align}\label{eq: error}
    E(w) = \frac{1}{2}Q(\frac{\mu_1^Tw}{\sqrt{w^T \Sigma_1 w}}) + \frac{1}{2}Q(-\frac{\mu_2^Tw}{ \sqrt{w^T \Sigma_2 w}})
\end{align}
where $Q(\cdot)$ is the integral of the tail of the standard normal distribution. 
\end{lemma}

The goal of this paper is to compute and characterize the generalization error of linear regression using different regularizers for the linear binary classifier with Gaussian mixture model. As can be seen from Lemma \ref{lem:gen_error}, this requires us to characterize the four quantities
\[ \mu_1^Tw~~,~~\mu_2^Tw~~,~~w^T\Sigma_1w~~,~~w^T\Sigma_2w \]
In fact, in much of the subsequent analysis, we shall assume $\Sigma_1 = \Sigma_2 = \sigma^2I$, which implies we need to characterize only the following three quantities
\begin{equation}\label{eq:inner-prod}
    \mu_1^Tw~~,~~\mu_2^Tw~~,~~\|w\|^2
\end{equation}
Since the data model that we are considering is a Gaussian mixture, we shall make use of the Convex Gaussian Min-Max Theorem (CGMT) (\cite{thrampoulidis2015regularized}), which is a tight and extended version of a classical Gaussian comparison inequality (\cite{gordon1985some}). 

\subsection{Convex Gaussian Min-Max Theorem}

The CGMT framework has been developed to analyze the properties of the solutions to non-smooth regularized convex optimization problems and has been successfully applied to characterize the precise performance in numerous applications such as $M$-estimators, generalized lasso, massive MIMO, phase retrieval, regularized logistic regression, adversarial training, max-margin classifiers, distributionally robust regression, and others (see \cite{stojnic2013framework,thrampoulidis2018precise,salehi2019impact,thrampoulidis2015lasso,abbasi2019performance,salehi2018precise,miolane2021distribution,taheri2021fundamental, aubin2020generalization,javanmard2022precise,montanari2019generalization,salehi2020performance,aolaritei2023}. In this framework, a given so-called primary optimization $\textbf{(PO)}$ problem, is associated with a simplified auxiliary optimization $\textbf{(AO)}$ problem from which properties of the optimal solution can be tightly inferred. Specifically, the $\textbf{(PO)}$ and $\textbf{(AO)}$ problems are defined as follows:
\begin{align}
\Phi(\mathbf{G})&:=\min _{\mathbf{w} \in \mathcal{S}_{\mathbf{w}}} \max _{\mathbf{u} \in \mathcal{S}_{\mathbf{u}}} \mathbf{u}^{\top} \mathbf{G w}+\psi(\mathbf{w}, \mathbf{u}) \!\! &\!\!{\textbf{(PO)}} \nonumber\\
\phi(\mathbf{g}, \mathbf{h}) & \!:=\!\!\min _{\mathbf{w} \in \mathcal{S}_{\mathbf{w}}} \max _{\mathbf{u} \in \mathcal{S}_{\mathbf{u}}}\|\mathbf{w}\|_2 \mathbf{g}^{\top} \mathbf{u}\!+\!\|\mathbf{u}\|_2 \mathbf{h}^{\top} \mathbf{w}\!+\!\psi(\mathbf{w}, \mathbf{u}) \!\! &\!\!{\textbf{(AO)}} \nonumber
\end{align}
where $\mathbf{G} \in \mathbb{R}^{m \times n}, \mathbf{g} \in \mathbb{R}^m, \mathbf{h} \in \mathbb{R}^n, \mathcal{S}_{\mathbf{w}} \subset \mathbb{R}^n, \mathcal{S}_{\mathbf{u}} \subset \mathbb{R}^m$ and $\psi: \mathbb{R}^n \times \mathbb{R}^m \rightarrow \mathbb{R}$. Denoting any optimal minimizers of $\textbf{(PO)}$ and $\textbf{(AO)}$ as $\mathbf{w}_{\Phi}:=\mathbf{w}_{\Phi}(\mathbf{G})$ and $\mathbf{w}_\phi:=\mathbf{w}_\phi(\mathbf{g}, \mathbf{h})$, respectively, CGMT states:

\begin{theorem}[CGMT~\cite{thrampoulidis2018precise}] \label{thm:cgmt} Let $\mathcal{S}_{\mathbf{w}}$, $\mathcal{S}_{\mathbf{u}}$ be convex compact sets, $\psi$ be continuous and convex-concave on $\mathcal{S}_{\mathbf{w}} \times \mathcal{S}_{\mathbf{u}}$, and, $\mathbf{G}, \mathbf{g}$ and $\mathbf{h}$ all have entries iid standard normal. Let $\mathcal{S}$ be an arbitrary open subset of $\mathcal{S}_{\mathbf{w}}$ and $\mathcal{S}^c := \mathcal{S}_{\mathbf{w}} / \mathcal{S}$. Denote by $\Phi_{\mathcal{S}^c}(\mathbf{G}) $ and $\phi_{\mathcal{S}^c}(\mathbf{g}, \mathbf{h}) $ the optimal costs of $\textbf{(PO)}$ and $\textbf{(AO)}$ respectively when $\mathbf{w}$ is minimized over  $\mathcal{S}^c$. If there exist constants $\bar{\phi} < \bar{\phi}_{\mathcal{S}^c}$ such that $\phi(\mathbf{g}, \mathbf{h}) \stackrel{p}{\longrightarrow} \bar{\phi}$, and $\phi_{\mathcal{S}^c}(\mathbf{g}, \mathbf{h}) \stackrel{p}{\longrightarrow} \bar{\phi}_{\mathcal{S}^c}$, (converge in probability), then $\lim _{n \rightarrow \infty} \mathbb{P}\left(\mathbf{w}_{\Phi}(\mathbf{G}) \in \mathcal{S}\right)=1$.

\end{theorem}

Let $f:\mathbb{R}^{d} \rightarrow \mathbb{R}$ be a convex function. For $w \in \mathbb{R}^{d}$ and $t > 0$ the Moreau envelope of function $f$ (\cite{moreau1965proximite}) is defined as
\begin{align*}
    \textbf{e}_f (w; t) := \min_{x} \frac{1}{2t} \|w-x\|^2 + f(x)
\end{align*}

We call a convex $f$ separable, when $f(w) = \sum_{i=1}^d f_i(w_i)$ for convex $f_i: \mathbb{R} \rightarrow \mathbb{R}$.

\subsection{Optimal solutions with oracle access to \texorpdfstring{$\mu_i$}{}}

We need to introduce the following two definitions before proceeding further:

\begin{definition}
    We say that $w \in \mathbb{R}^d$ is $k$-sparse if it has at most $k$ non-zero entries
\end{definition}

\begin{definition}
    We say that $w \in \mathbb{R}^d$ is a $1$-bit vector if each of its entries is equal to $\pm a$, for some $a\in \mathbb{R}$
\end{definition}
In the over-parametrized regime, since $n<d$, it is not possible to reliably estimate the means $\mu_1$ and $\mu_2$ (this is further confounded when the labels have errors). Nonetheless, it is useful to see what the optimal linear classifiers would look like when one has oracle access to the means. We will do this for the general case, as well as for the $k$-sparse and $1$-bit classifiers. 

\begin{lemma}\label{lem:optimal}

    1. The overall optimal classifier is $w = \mu_1 -\mu_2$, which has performance $Q\left(\sqrt{\frac{d(1-r)}{2\sigma^2}}\right)$
    
    2. The optimal $1$-bit classifier is $w = sign(\mu_1 - \mu_2)$, which has performance $Q\left(\sqrt{\frac{d(1-r)}{\pi\sigma^2}}\right)$

    3. The optimal $k$-sparse classifier is obtained from taking the $k$ coordinates of $\mu_1 - \mu_2$ with the largest magnitude and zeroing out the rest 
\end{lemma}

\subsection{Regularized linear regression in the presence of corruption in labels}\label{subs: setting}

It is natural to use regularized linear regression for classification when not all labels are reliable. As mentioned earlier, $c<0.5$ stands for the corruption rate, meaning that $\frac{cn}{2}$ labels within each class are corrupt. Let $f(\cdot)$ be an arbitrary convex regularizer and $\lambda$ be the regularization strength. The present paper is concerned with analyzing the case of the linear regression applied to GMM (cf. \ref{bin_class}) with means $\mu_1$ and $\mu_2$ and covariance $\sigma^2I$.   After reordering the training data so that the first $\frac{n}{2}$ columns of the data matrix correspond to the points from the first class, our analysis reduces to the following optimization problem: 
\begin{align}\label{eq:est}
    \min_w \| (A+M) w -z  \|_2^2 + \lambda  f(w)
\end{align}
Where the entries of $A \in \mathbb{R}^{n \times d}$  are i.i.d. $\mathcal{N}(0,\sigma^2)$, $M^T = 
    ( \underbrace{\begin{matrix}
       \mu_1 & ... &
    \mu_1
    \end{matrix}}_{\frac{n}{2}} \underbrace{\begin{matrix}
       \mu_2 & ... & 
    \mu_2 
    \end{matrix}}_{\frac{n}{2}})$ encodes the means corresponding to each class, $z^T = \begin{pmatrix}
        \mathds{1}^T_{\frac{(1-c)n}{2}} & -\mathds{1}^T_{\frac{n}{2}} &
        \mathds{1}^T_{\frac{cn}{2}}
    \end{pmatrix}$ encodes the training labels.

\subsection{Approach and intuition behind it}\label{subs: overview_and_intuition}

As mentioned earlier, we are interested in computing the inner products and the norm (\ref{eq:inner-prod}). We focus on the over-parametrized high-dimensional regime where $\frac{d}{n} = \gamma>1$ is fixed and $n\rightarrow\infty$.  In this regime, we show that the quantities in (\ref{eq:inner-prod}) concentrate and we determine their asymptotic values. Moreover, in the cases of $\ell_1$ and $\ell_\infty$ regularization, we do the same for the sparsity and compression rates, respectively. 

To do so, we will employ the CGMT framework. This starts by using the Fenchel dual of the quadratic loss to rewrite (\ref{eq:est}) in the following min-max form:
\begin{align*}
    \min_w \max_v   v^TAw + v^TMw - v^Tz - \frac{1}{4}\|v\|^2 + \lambda  f(w) 
\end{align*}
Applying CGMT and explicitly performing the maximization over $v$ in the $\textbf{(AO)}$, we obtain: 
\begin{align}\label{eq: ao_explicit}
    &\min_w \quad\ (g^T w + \sqrt{n} \sigma \|w\|)_{\ge 0 }^2 + \frac{n}{2} (\Tilde{\mu_1}^T w - (1-2c))^2 +\frac{n}{2} (\Tilde{\mu_2}^T w + (1-2c))^2 + n (\frac{1}{\sqrt{n}}a^T w - 2 \sqrt{c(1-c)})^2 + \lambda f(w)
\end{align}
where we have used the following notation
\begin{align*}
    g, a, a_1, a_2 \in \mathcal{N}(0,\sigma^2I) \\
     \tilde{\mu_i} = \mu_i + \sqrt{\frac{2}{n}}a_i \quad  i \in \{1,2\} \\
     x_{\ge 0} = \max{(x, 0)}
\end{align*}
It is insightful to examine each term in (\ref{eq: ao_explicit}) as it provides some explanation of the phenomenona observed in the simulations. 

$\bullet $ The sum of the second and the third terms  $\frac{n}{2} (\Tilde{\mu_1}^T w - (1-2c))^2 +\frac{n}{2} (\Tilde{\mu_2}^T w + (1-2c))^2 $ encourages the minimizer $w$ to align with $\mu_1 - \mu_2 + \sqrt{\frac{2}{n}}(a_1-a_2)$. As seen earlier, the oracle-based optimal $w$ is $\mu_1 - \mu_2$. Thus, it is these two terms that encourage $w$ to approach the oracle-based optimal. Note that the extra term $\sqrt{\frac{2}{n}}(a_1-a_2)$ has $\frac{2\sigma^2}{n(1-r)}$ of the variance of $\mu_1-\mu_2$ and represents the effects of the variance in the training data from the GMM. The second and third terms represent fitting the correctly labeled data. 

$\bullet$ The fourth term $n(\frac{1}{\sqrt{n}}a^T w - 2 \sqrt{c(1-c)})^2$ encourages $w$ to align with the arbitrary Gaussian vector $\frac{a}{\sqrt{n}}$. This represents the minimizer's attempt to interpolate the corrupted labels, as it vanishes when $c=0$. 

$\bullet$ The first term $(g^T w + \sqrt{n} \sigma \|w\|)_{\ge 0 }^2$ makes $w$ partly align with a random direction $-g$ and is independent of $c$. It represents the fact that the problem is over-parametrized. 

$\bullet$ The last term $\lambda f(w)$ is the regularization term. 

Ideally, we would like the minimizer to minimize the second and third terms and to ignore the first and fourth ones. We now provide some intuition as to why the regularization term $f(w)$ helps with this. Due to the equivalence of norms, our argument applies to any norm $f(w) = \|w\|_p$, and so, for simplicity, we shall focus on $p=2$ and $f(w) = \|w\|_p^2$. 

Due to the existence of the $\|w\|_2$ regularizer, the minimizer would like to minimize the terms in (\ref{eq: ao_explicit}) with as small a $\|w\|^2$ as possible. This is much easier to do for the second and third terms, since the squared norm of the vectors ${\tilde \mu}_1$ and ${\tilde \mu}_2$ is $d(1+\frac{2\sigma^2}{n})$, than it is for the fourth term where the squared norm of the vector $\frac{1}{\sqrt{n}}a$ is $\frac{\sigma^2d}{n}$. Thus, the minimizer will reduce the second and third terms at the expense of the fourth one. In other words, the regularization
term encourages the regressor to interpolate the correctly labeled data, at the expense of the corrupted labels, which improves performance and is what we hoped it would do. In fact, the larger $\lambda$ is, the stronger the incentive to ignore the mislabeled data. 
\section{MAIN RESULTS}
In this section we provide rigorous results along some remarks which will help to gain more insights on the general problem.
\subsection{Precise results}

The reader can find a theorem describing the generalization error for an arbitrary convex regularizer $f$ below.
\vspace{3mm}
\begin{theorem}[Master theorem]\label{thm: AO_gen}
The generalization error resulting from the application of the linear regression with a separable convex regularizer $f = \sum_{i=1}^d f_i(w_i)$, with $f_i$'s being identical and regularization strength $\lambda$ to the Gaussian mixture model with means $\mu_1, \mu_2$ and covariance $\sigma ^ 2 I$ is equal to 
$$E(f, \lambda) = Q\left(\frac{\gamma}{2\sqrt{\frac{1}{n\tau ^ 2} - (1-c)(\frac{\gamma}{2} - 1) ^ 2 - c (\frac{\gamma}{2} + 1) ^ 2}}\right)$$ 
for $\tau$ and $\gamma$ defined by the scalar optimization
\begin{align*}
     &\min_{\tau \geq 0} \max_{\beta \geq 0, \gamma}  \frac{\beta\tau}{2} ( - \frac{n}{4} \gamma^2 - \frac{nd (1-r)}{2\sigma^2} (\frac{\gamma}{2} - 1 + 2c)^2  + n) + \frac{\beta }{2\tau} (1 - \frac{d}{n}) - \frac{1}{4}\beta^2 + d \lambda \mathbb{E } e_{ f_i} (\Xi(\gamma, \tau) G; \frac{\lambda}{\beta\tau n \sigma^2})
\end{align*}
where the expectation is taken over $G \sim \mathcal{N}(0,1)$ and  $\Xi(\gamma, \tau) := \sqrt{\frac{1-r}{2\sigma^4}(\frac{\gamma}{2} -1+2c)^2  + \frac{1}{n^2 \tau^2 \sigma^2}}$
\end{theorem}
Next, we proceed to provide some specialized results for prominent regularizers. In the case of $f=\|.\|_2$, it turns out that $\hat{w}$ from Theorem \ref{thm: AO_gen} takes form of $w(\alpha, \gamma) = \alpha g + \gamma(\mu_1 - \mu_2)$ for some $\alpha, \gamma \in \mathbb{R}$ (see Appendix). This allows for an easier way of characterising the desired generalization error than the one suggested by Theorem \ref{thm: AO_gen}. 
\vspace{3mm}
\begin{theorem}[$\ell_2$ regularization] \label{thm: ridge}

The generalization error resulting from the application of the ridge regression with regularization strength $\lambda$ to the Gaussian mixture model with means $\mu_1, \mu_2$ and covariance $\sigma ^ 2 I$ is equal to 

$$E(\ell_2, \lambda) = Q\left(\frac{\gamma d(1 - r)}{\sqrt{\alpha ^ 2\sigma ^ 4 d + 2\sigma ^ 2  \gamma ^ 2 d(1 - r)}}\right)$$ 

where $\alpha, \gamma$ are
defined by the following two-dimensional convex optimization problem:
$$ \min_{\alpha, \gamma} (\alpha\sigma^2d + \sqrt{n \sigma ^ 2 \Omega(\alpha, \gamma) + \Theta(\alpha, \gamma)})_{\ge 0} ^ 2 + \lambda \Omega(\alpha, \gamma)$$
with
 \begin{align*}
\Omega(\alpha, \gamma)  &=   \alpha ^ 2 \sigma^2 d + 2d(1-r)\gamma ^ 2 \\
\Theta(\alpha, \gamma)  &=   n(1-c)(d(1-r)\gamma - 1) ^ 2 + nc(d(1-r)\gamma + 1) ^ 2   
\end{align*}

\end{theorem}
For $\ell_1$ regularization, one can find the sparsity rate in addition to analyzing performance. 
\vspace{3mm}
\begin{theorem}[$\ell_1$ regularization] \label{thm: AO_l1}
The generalization error resulting from the application of the $\ell_1$ regularized regression with regularization strength $\lambda$ to the Gaussian mixture model with means $\mu_1, \mu_2$ and covariance $\sigma ^ 2 I$ is equal to
$$E(\ell_1, \lambda) = Q\left(\frac{\gamma}{2\sqrt{\frac{1}{n\tau ^ 2} - (1-c)(\frac{\gamma}{2} - 1) ^ 2 - c (\frac{\gamma}{2} + 1) ^ 2}}\right)$$ 
for $\beta, \tau$ and $\gamma$ defined by the optimization: 
\begin{align*}
    &\min_{\tau \geq 0} \max_{\beta \geq 0, \gamma} -\frac{2d\beta}{n\tau} Q(\frac{\lambda}{n \beta \tau \tilde{\sigma}}) + \frac{\beta }{2\tau} - \frac{1}{4}\beta^2 + \frac{\beta\tau n}{2}[d\sigma^2(\frac{\lambda}{n\beta\tau\sigma^2})^2(2(s^2+1)Q(\frac{1}{s}) - \frac{2s}{\sqrt{2\pi}}e^{-\frac{1}{2s^2}}) + \\
    &+ \frac{2\gamma d}{\sigma^2}Q(\frac{\lambda}{n \beta \tau \tilde{\sigma}}) (\frac{1-2c}{2} - \frac{\gamma}{4})(1-r) + \frac{\gamma^2}{4} +  1 - (1-2c)\gamma] + \frac{2d\lambda\tilde{\sigma}}{\sigma^2\sqrt{2\pi}} e^{-\frac{\lambda^2}{2(n\beta\tau\tilde{\sigma})^2}} -  \frac{2d\lambda^2}{n\beta\tau \sigma^2} Q(\frac{\lambda}{n\beta\tau\tilde{\sigma}})
\end{align*}
where
\begin{align*}
\tilde{\sigma}^2 & = \frac{\sigma^2}{n^2\tau^2} + 2(\frac{\gamma}{4} - \frac{1-2c}{2})^2(1-r) \text{ and }s  = \frac{\tilde{\sigma}n\beta\tau}{\lambda}
\end{align*}
Furthermore, the corresponding solution is $\lceil{2Q(\frac{\lambda}{n\beta\tau\tilde{\sigma}})}\rceil$-sparse
\end{theorem}

Finally, in the case of $f = \ell_\infty$ we were also able to calculate the compression rate.

\vspace{3mm}

\begin{theorem}[$\ell_{\infty}$ regularization] \label{thm: AO_linf}
The generalization error resulting from the application of the $\ell_\infty$ regularized regression with regularization strength $\lambda$ to the Gaussian mixture model with means $\mu_1, \mu_2$ and covariance $\sigma ^ 2 I$ is
$$E(\ell_\infty, \lambda) = Q\left(\frac{\gamma}{2\sqrt{\frac{1}{n\tau ^ 2} - (1-c)(\frac{\gamma}{2} - 1) ^ 2 - c (\frac{\gamma}{2} + 1) ^ 2}}\right)$$ 
for $\beta, \tau$ and $\gamma$ defined by the optimization: 
\begin{align*}
     &\min_{\tau, \delta \geq 0} \max_{\gamma} \delta + \bigg( \frac{1 }{2\tau} + \frac{\tau}{2} ( - \frac{n}{4} \gamma^2 + n ) - \frac{n d \sigma^2\tau \Xi^2}{2}-  \frac{n d \sigma^2\tau \delta \Xi}{\lambda\sqrt{2\pi}} exp(- \frac{\delta^2}{2\lambda^2\Xi^2}) + n d \sigma^2\tau ( \Xi^2 + \frac{\delta^2}{\lambda^2}) Q (\frac{\delta}{\lambda \Xi}) \bigg)^2_{\geq 0} 
\end{align*}
where 
$$\Xi(\gamma, \tau) := \sqrt{\frac{1-r}{2\sigma^4}(\frac{\gamma}{2} -1+2c)^2  + \frac{1}{n^2 \tau^2 \sigma^2}}$$
Moreover, $k$ of the weights are equal to $\frac{\delta}{\lambda}$,  $k$ of the weights are equal to $-\frac{\delta}{\lambda}$, and all others weights lie in between, where $k = \lfloor{d Q(\frac{\delta}{\lambda\Xi})}\rfloor $
\end{theorem}

\subsection{Explicit approximations for large \texorpdfstring{$\lambda$}{}}\label{sec: approx}

While the theorems above accurately describe the generalization errors, they are not as explicit as one might wish. To overcome this, we show for $\lambda$ large enough, the first term of (\ref{eq: ao_explicit}) is negligible compared to $\lambda f(w)$.  This suggests dropping that term completely from the $\textbf{(AO)}$. Leaving the technical details for the Appendix, we summarize the implications of this approximation in the remarks below. In Section \ref{sec:numerics} we will observe that these approximations do work well when $\lambda$ is large. 

\begin{remark}\label{rem: l2closed}
The following approximation for the generalization error takes place in the case of the ridge regression if $\lambda \gg \sigma^2n$: 
\begin{align*}
    \gamma = \frac{1-2c}{d(1-r + \frac{2\sigma^2}{n}) + 2\frac{\lambda}{n}}
\end{align*}
\begin{align*}
    \beta = \frac{2\sqrt{nc(1-c)}}{d \sigma^2 + \lambda}
\end{align*}
\begin{align*}
    E(\ell_2, \lambda) = Q \left(\frac{d(1-r) \gamma }{\sigma \sqrt{2 d (1-r+\frac{2\sigma^2}{n}) \gamma^2 + d \sigma^2 \beta ^2}}\right)  
\end{align*}
\end{remark}
\begin{remark}\label{rem: l2Qfunc}
    Assume that $\sigma \ll \sqrt{n}$. Then the following approximation can be made: $$\frac{d(1-r) \gamma }{\sigma \sqrt{2 d (1-r+\frac{2\sigma^2}{n}) \gamma^2 + d \sigma^2 \beta ^2}} \approx \frac{\sqrt{d(1-r)} }{\sigma \sqrt{2}}$$ 
\end{remark}
Note that the $Q$-function applied to the argument above is a negligibly small number for big enough $d$. Thus, informally, this remark can be stated as follows:  if the problem is high-dimensional and $\sigma$ is small enough, the negative effects of any corruption rate $c < 0.5$ can be completely eliminated via ridge regression with a sufficiently large regularization strength.  
\begin{remark}\label{rem:l_1_sparse}
    Dropping the first term of (\ref{eq: ao_explicit}) in the case of the $\ell_1$-regularized regression and assuming that $\lambda$ is large enough, one can show that $w_i = 0$ unless $|t_i|$ is close to $\lambda$, where $t \in \mathbb{R} ^ d$  is defined via
    $$t =  \frac{n}{2} \gamma(\Tilde{\mu_1} - \Tilde{\mu_2}) + \sqrt{n} \beta a$$
    Moreover, it turns out that the optimal scalars $\gamma$, $\alpha$, and $\beta$ are such that $|t_i| \le \lambda$ holds with high probability. This suggests that $\ell_1$ regularization tries to kill all the components of $w$ apart from the ones corresponding to the top entries of $|\gamma(\Tilde{\mu_1} - \Tilde{\mu_2}) + \sqrt{n} \beta a|$ where the latter can be regarded as an approximation to $\mu_1 - \mu_2$. This is similar to the description of the optimal sparse classifier from Lemma \ref{lem:optimal}. Thus, for large enough $\lambda$, the $\ell_1$ regularizer tries to find as sparse a solution as possible that aligns itself with the top entries of $\mu_1-\mu_2$.  
\end{remark}
\begin{remark}\label{rem:l_inf_compr} Assuming that $\lambda$ is large enough, the following approximation can be made for the $w$ found from the $\ell_\infty$-regularized regression:   
 $$w_i = - \frac{\delta}{\lambda} \mbox{sign}(( \frac{n}{2} \gamma(\Tilde{\mu_1} - \Tilde{\mu_2}) + \sqrt{n} \beta a)_i)$$
 Where $\gamma, \beta$ and $\delta$ are defined by the optimization:
 \begin{align*}
    &\min_{\delta \geq 0 } \max_{\gamma, \beta \ge 0} \delta -n(1-2c) \gamma -\frac{\gamma^2}{2} - 2n  \sqrt{c(1-c)} \beta  -\frac{\beta^2}{4} - \frac{\delta}{\lambda} \| \frac{n}{2}\gamma(\Tilde{\mu_1} - \Tilde{\mu_2}) + \sqrt{n} \beta a\|_1
\end{align*}
for $a \sim \mathcal{N}(0, \sigma^2 I)$

Since all entries of $w$ have the same magnitude and since (\ref{eq: error}) is unaffected by a scaling of $w$, this implies that we can replace $w$ by $sign(w)$ without any loss of performance. Thus each component of the optimal $w$ can be encoded by a single bit. 
\end{remark}
\section{A COMPRESSION SCHEME}\label{sec: add_compress}

As discussed in Remark \ref{rem:l_inf_compr}, $\ell_\infty$-regularization with large $\lambda$ can be used for $1$-bit compression via $\tilde{w} = \mbox{sign}(w)$. One might wonder whether using the same compression scheme for an arbitrary $f$-regularized solution would still retain good performance. Turns out that this scheme does succeed for $f = \ell_2^2$ and $f = \ell_1$ in the small noise ($\sigma \ll \sqrt{n}$) and large $\lambda$ regime. For $f = \ell_1$, this is discussed in Remark \ref{rem:l_1_sparse}. For $f = \ell_2^2$, as discussed in Section \ref{subs: overview_and_intuition}, the corresponding $w$ strives to align with $\mu_1 - \mu_2$.  This makes $\mbox{sign}(w)$ approximate $\mbox{sign}(\mu_1 - \mu_2)$ which, according to Lemma $\ref{lem:optimal}$, is the optimal $1$-bit classifier. 
\section{NUMERICAL RESULTS}\label{sec:numerics}
To showcase our results, we performed extensive simulations using synthetic data produced according to the assumptions described in the previous sections. We generated points equiprobably from two distributions $\mathcal{N}(\mu_1, \sigma^2I)$ and $\mathcal{N}(\mu_2, \sigma^2I)$, where the corresponding components of the $\mu_i$ were iid standard normal with cross-correlation $-1\leq r\leq 1$, and with labels $+1$ and $-1$ respectively, a fraction $c$ of those which were subsequently corrupted. Making use of the MATLAB\textsuperscript{\texttrademark} CVX package (\cite{gb08,cvx}), we trained classifiers that minimize the $\ell_2$, $\ell_1$ and $\ell_\infty$ regression objectives for this data. We simulated the generalization error of these classifiers, and compared it to the expressions obtained in theorems \ref{thm: ridge}, \ref{thm: AO_l1}, and \ref{thm: AO_linf} (cf. Fig.\ref{ridgevary}, Fig.\ref{l1vary}, Fig.\ref{linfvary}). Moreover, we have also plotted the predicted sparsity and compression rate for the $\ell_1$ and $\ell_{\infty}$ cases. In the following figures, we took the nominal values $n = 200, d = 2000, c = 0.2, r = 0.8, \sigma = 2$ as a starting point. Then, to analyze the effects of these parameters independently, we vary exactly one parameter per experiment while keeping the others fixed. In the subsections below, we go over the plots and interpret their results.
\subsection{Ridge regression}
\begin{figure}
\centerline{\includegraphics[width=0.55\textwidth]{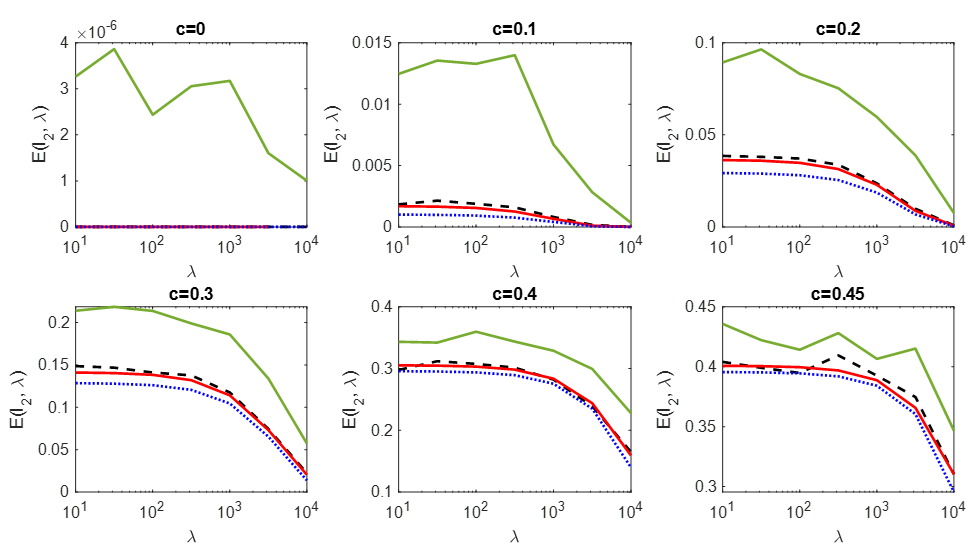}}
\centerline{\includegraphics[width=0.55\textwidth]{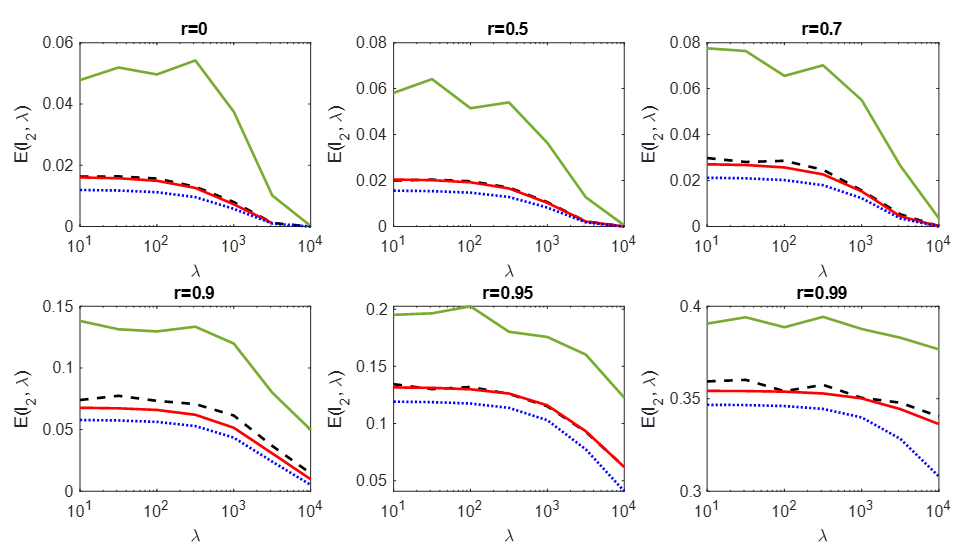}}
\centerline{\includegraphics[width=0.55\textwidth]{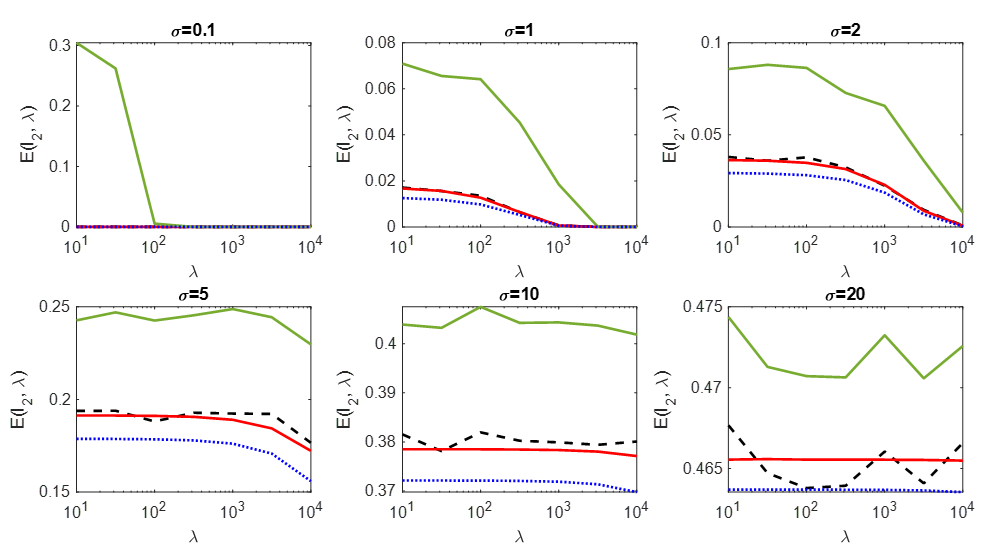}}
\centerline{\includegraphics[width=0.55\textwidth]{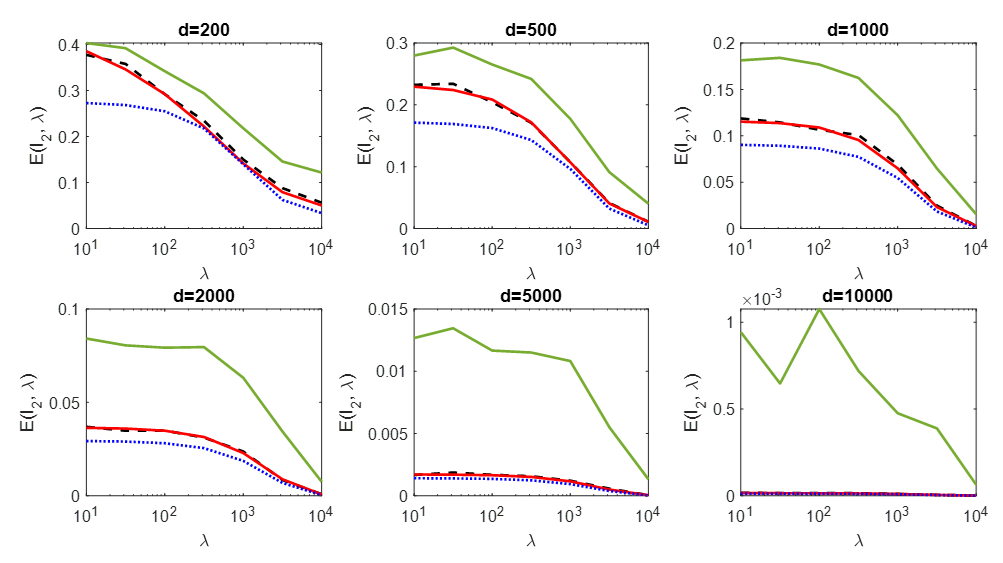}}
\caption{Black dashed line: simulated error. Red line: Theorem \ref{thm: ridge} prediction. Blue dotted line: closed-form approximation for large $\lambda$. Green line: true error for $sign(w)$. These 4 sets of plots characterize the generalization error of ridge regression when one parameter changes while the others are fixed.} 
\label{ridgevary}
\end{figure}

For the ridge regression, we examined the effects of $r,c, \sigma, \frac{d}{n} $ on the generalization error in Fig.~\ref{ridgevary}. As expected, the generalization error improves as the regularization strength increases. The $E(\ell_2, \lambda)$, simulated directly by solving ridge regression, and the prediction for it derived from Theorem \ref{thm: ridge} match very closely. Moreover, the closed-form approximation formulated in Section \ref{sec: approx} follows the true $E(\ell_2, \lambda)$ closely for large values of $\lambda$. In addition, it appears to be a lower bound for the generalization error for all $\lambda > 0$, though we leave proving this to future work. Finally, in light of Section \ref{sec: add_compress}, we also considered the sign of the solution as a possible classifier. It turns out one does not lose much in performance by compressing each weight to a single bit. Furthermore, it is evident that increasing either of $c, r, \sigma$ increases the classification error and, on the opposite, increasing $\frac{d}{n}$, decreases it. 
\subsection{\texorpdfstring{$\ell_1$}{} regularization}
\begin{figure} [ht]
\vspace{.3in}
\centerline{\includegraphics[width=0.60\textwidth]{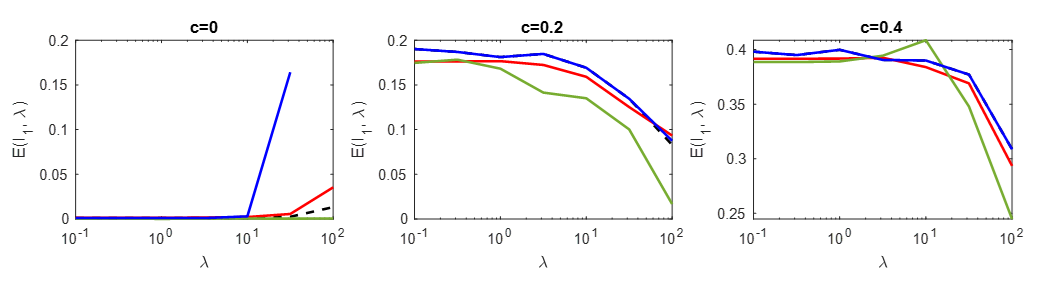}}
\centerline{\includegraphics[width=0.61\textwidth]{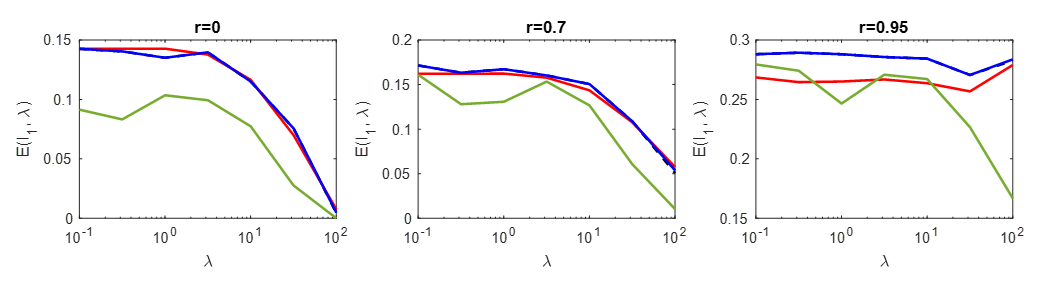}}
\centerline{\includegraphics[width=0.60\textwidth]{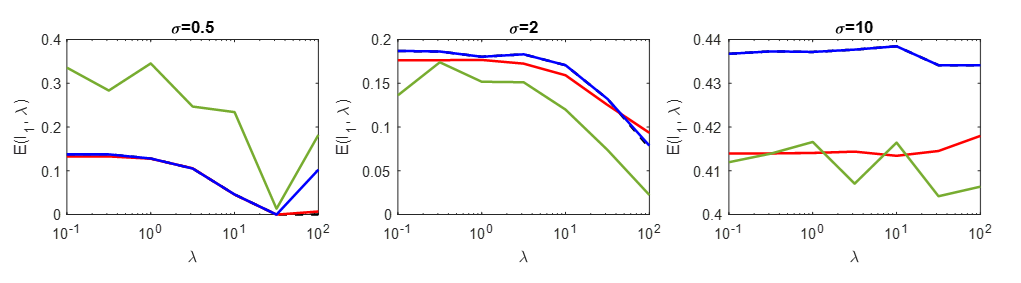}}
\caption{Black dashed line: simulated error. Red line: prediction from Theorem \ref{thm: AO_l1}.  Green line: $\mbox{sign}(w)$. Blue line: sparsified solution}
\label{l1vary}
\end{figure}

\begin{figure}[ht]
\vspace{.3in}
\centerline{\includegraphics[width=0.40\textwidth]{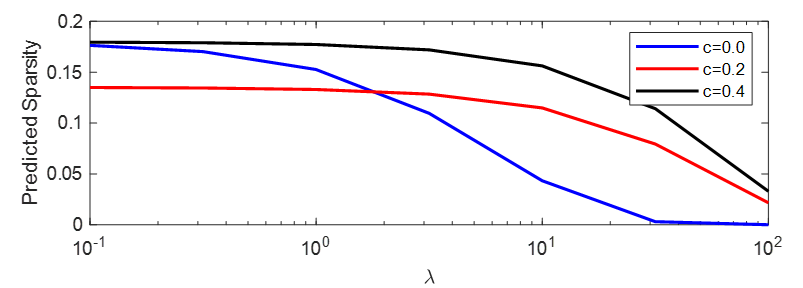}}
\centerline{\includegraphics[width=0.40\textwidth]{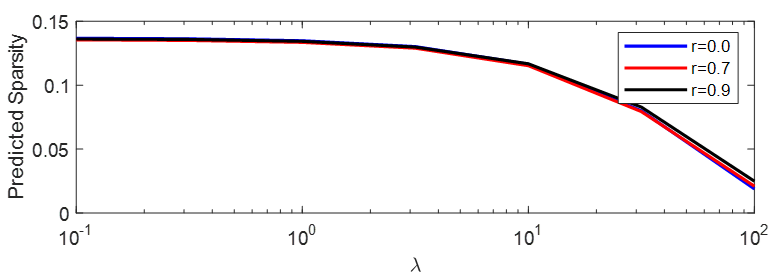}}
\centerline{\includegraphics[width=0.40\textwidth]{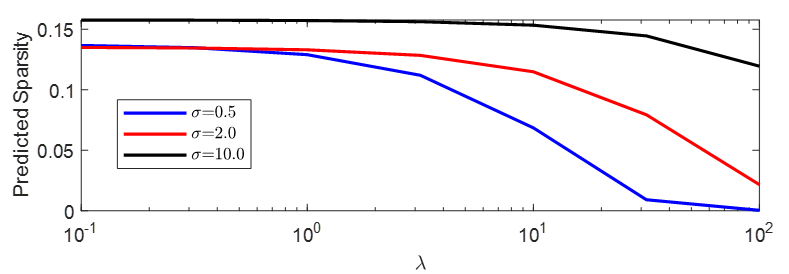}}
\caption{Effect of each parameter on the sparsity rate}
\label{sparvary}
\end{figure}

Similar to the previous section, in Fig.~\ref{l1vary} we compared the simulated error to the error predicted by Theorem \ref{thm: AO_l1} and observed a close match. The solution has many entries that are very close to zero, but are not exactly equal to it because of small-scale numerical inaccuracies, and so we sparsified this solution according to the rate suggested by Theorem \ref{thm: AO_l1}. We observed that this does affect the performance. Finally, we also compress the $\ell_1$ solution down to $1$ bit. The effects of changing $c, r, \sigma$ on the generalization error are similar to what was observed in the ridge regression case. Interestingly, changing these parameters has different impacts on the theoretical prediction for the sparsity rate. Fig.~\ref{sparvary} proposes the sparsity rate does not change much with $r$, whereas increasing $\sigma, c$ leads to difficulty in sparsifying. Also note that we obtain sparsity rates close to zero as we increase $\lambda$. Pairing the observations from Figs.~\ref{l1vary} and \ref{sparvary}, we see that one can attain a good generalization error with an extremely sparse estimator. 
For instance, when $\sigma = 0.5$, we get an $18$-sparse estimator (in an ambient dimension of 2000) with a generalization error of as little as $0.0003$. This is despite the fact that the underlying GMM model had no inherent sparsity structure. 
\subsection{\texorpdfstring{$\ell_\infty$}{} regularization}
\begin{figure}[ht]
\centerline{\includegraphics[width=0.60\textwidth]{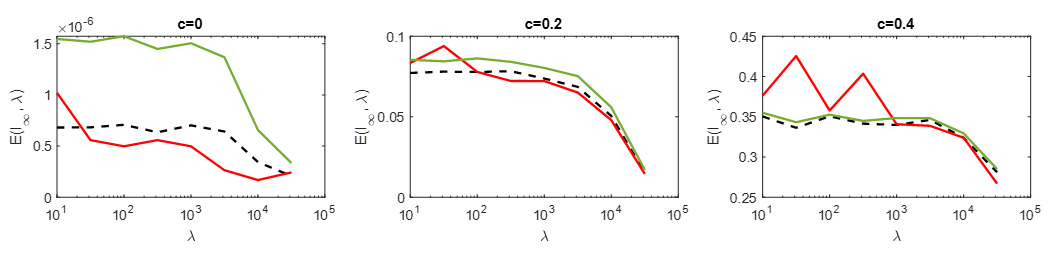}}
\centerline{\includegraphics[width=0.60\textwidth]{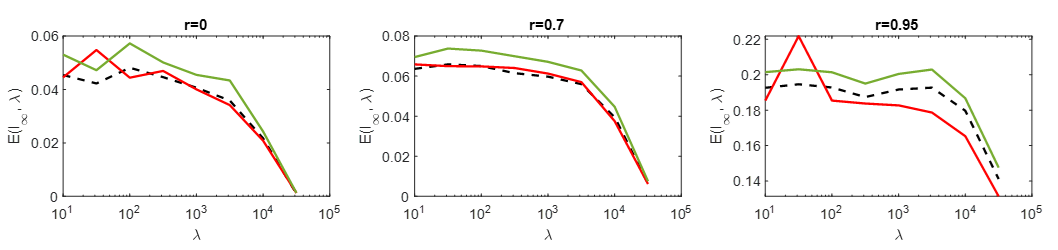}}
\centerline{\includegraphics[width=0.60\textwidth]{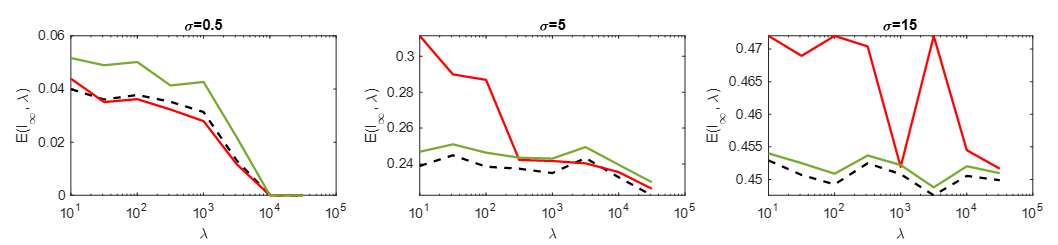}}
\caption{Black dashed line: $\ell_{\infty}$ error, Red line: AO prediction, Green line: $sign(w)$}
\label{linfvary}
\end{figure}

\begin{figure}[ht]
\vspace{.3in}
\centerline{\includegraphics[width=0.40\textwidth]{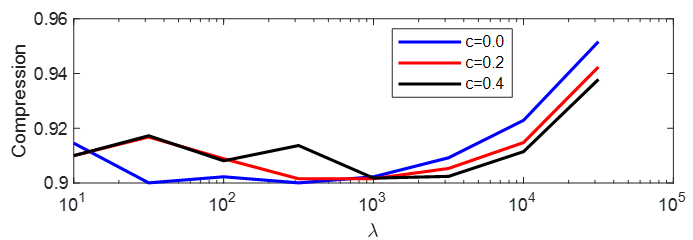}}
\centerline{\includegraphics[width=0.40\textwidth]{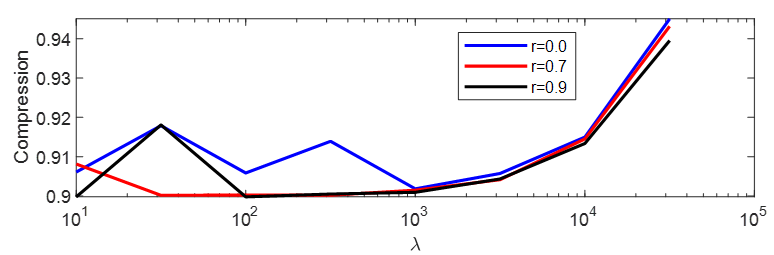}}
\centerline{\includegraphics[width=0.40\textwidth]{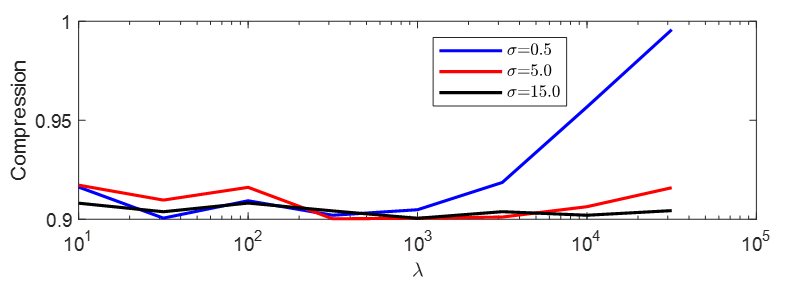}}
\caption{Effect of each parameter on the compression rate}
\label{compvary}
\end{figure}

Repeating the same procedure as before, we verified that the predicted error matches the simulated error, as can be seen in Fig.~\ref{linfvary}. Moreover, we checked the same for the compression rate (cf. Fig.~\ref{compvary}). It is noteworthy that the $1$-bit estimator follows the simulated error closely. Analyzing the compression rate, it is evident that no matter how corrupt the labels are (larger $c$), one can still take the one-bit estimator and not lose out much on performance. The same could also be said about the angle between the means represented by $r$.

\section{CONCLUSION}

In this work we considered the problem of binary classification with corrupt labels through the lens of regularized linear regression. We used CGMT to derive sharp results for the generalization error for a general convex separable regularizer $f$. Both theoretically, and through simulations, we showed that regularization helps the generalization performance. We calculated the sparsity rate for the case $f = \ell_1$ and the compression rate for the case $f = \ell_\infty$ and validated the theoretical findings numerically. These evaluations suggest that good sparse classifiers and good one-bit classifiers often exist, even when the underlying GMM has no inherent sparsity structure, and even in the presence of high rates of label corruption. Possible future directions include extending this work to the case of multi-class classification and to other regularizers and loss functions. 



\clearpage

\bibliography{main}

\newpage

\appendix
\section{USEFUL TECHNICAL RESULTS}

The following results will be of use for proving the theorems. We will invoke them multiple times without referring to them explicitly:

{\bf An expectation formula. }{\it Let $X \sim \mathcal{N}(0, \sigma^2)$. Then $\mathbb{E}[(|X|-1)^2 \mathds{1}_{|X| > 1} ] = 2(\sigma^2+1)Q(\frac{1}{\sigma}) - \frac{2\sigma}{\sqrt{2\pi}}e^{-\frac{1}{2\sigma^2}} $.}

\begin{proof} 

Denote $Y = \frac{X}{\sigma} \sim \mathcal{N}(0,1)$.

$$\mathbb{E}[(|X|-1)^2 \mathds{1}_{|X| > 1}] = \mathbb{E}[X^2 \mathds{1}_{|X| > 1}] - 2\mathbb{E}[|X| \mathds{1}_{|X| > 1}] + \mathbb{E}[\mathds{1}_{|X| > 1}] = 2\mathbb{E}[X^2 \mathds{1}_{X > 1}] - 4\mathbb{E}[X \mathds{1}_{X > 1}] + 2\mathbb{E}[\mathds{1}_{X > 1}] = $$
$$= 2\sigma^2\mathbb{E}[Y^2 \mathds{1}_{Y > \frac{1}{\sigma}}] - 4 \sigma\mathbb{E}[Y\mathds{1}_{Y > \frac{1}{\sigma}}]+ 2\mathbb{E}[\mathds{1}_{Y > \frac{1}{\sigma}}] = \frac{2\sigma^2}{\sqrt{2\pi}} \int^{+\infty}_{\frac{1}{\sigma}}t^2 e^{-\frac{t^2}{2}} dt - \frac{4\sigma}{\sqrt{2\pi}}\int^{+ \infty}_{\frac{1}{\sigma}} t e^{-\frac{t^2}{2}} dt + 2Q(\frac{1}{\sigma}) =$$
$$= -\frac{2\sigma^2}{\sqrt{2\pi}} \int^{+\infty}_{\frac{1}{\sigma}}t d e^{-\frac{t^2}{2}}  - \frac{4\sigma}{\sqrt{2\pi}}\int^{+ \infty}_{\frac{1}{\sigma}} e^{-\frac{t^2}{2}} d\frac{t^2}{2} + 2Q(\frac{1}{\sigma}) = - \frac{2\sigma^2}{\sqrt{2\pi}} e^{-\frac{t^2}{2}}t _{|_{\frac{1}{\sigma}}^{+\infty}}  + \frac{2\sigma^2}{\sqrt{2\pi}} \int^{+\infty}_{\frac{1}{\sigma}} e^{-\frac{t^2}{2}} dt -$$
$$- \frac{4\sigma}{\sqrt{2\pi}}\int^{+ \infty}_{\frac{1}{2\sigma^2}} e^{-z} dz + 2Q(\frac{1}{\sigma}) = \frac{2\sigma}{\sqrt{2\pi}}e^{-\frac{1}{2\sigma^2}} + 2\sigma^2Q(\frac{1}{\sigma}) - \frac{4\sigma}{\sqrt{2\pi}}e^{-\frac{1}{2\sigma^2}} + 2Q(\frac{1}{\sigma}) =$$
$$ = 2(\sigma^2+1)Q(\frac{1}{\sigma}) - \frac{2\sigma}{\sqrt{2\pi}}e^{-\frac{1}{2\sigma^2}}  $$
\end{proof}

{\bf A square root trick. }{\it The following equality holds for any $x > 0$:
    $$ \sqrt{x} = \min_{\beta > 0} \frac{1}{2\beta} + \frac{\beta x}{2} $$}

\begin{proof}
    Differentiate the objective from the right hand side by $\beta$:

    $$ \frac{d}{d\beta}(\frac{1}{2\beta} + \frac{\beta x}{2}) = \frac{1}{-2\beta^2} + \frac{x}{2} $$

    We conclude that $\frac{1}{2\beta} + \frac{\beta x}{2}$ is minimized at $\beta = \frac{1}{\sqrt{x}}$. The value the objective takes at this point is 
    
$\frac{1}{2\beta} + \frac{\beta x}{2} = \frac{1}{\frac{2}{\sqrt{x}}} + \frac{\frac{1}{\sqrt{x}}x}{2} = \sqrt{x}$
\end{proof}

\section{PROOFS OF THE LEMMAS}
\subsection{Proof of Lemma \ref{lem:gen_error}} 
\begin{proof}
    By definition, 
    
    $$E(w) = \frac{1}{2}\mathbb{P}_{x \sim \mathcal{N}(\mu_1, \Sigma_1})(w^Tx < 0) + \frac{1}{2}\mathbb{P}_{x \sim \mathcal{N}(\mu_2, \Sigma_2)}(w^Tx \ge 0)$$

    Rewrite $x = \mu_1 + y_1$ for $x \sim \mathcal{N}(\mu_1, \Sigma_1)$ and $x = \mu_2 + y_2$ for $x \sim \mathcal{N}(\mu_2, \Sigma_2)$. Note that $y_1 \sim \mathcal{N}(0, \Sigma_1)$ and $y_2 \sim \mathcal{N}(0, \Sigma_2)$. We obtain:

    $$E(w) = \frac{1}{2}\mathbb{P}_{y_1 \sim \mathcal{N}(0, \Sigma_1)}(w^Ty_1 < -\mu_1^Tw) + \frac{1}{2}\mathbb{P}_{y_2 \sim \mathcal{N}(0, \Sigma_2)}(w^Ty_2 \ge - \mu_2^Tw)$$

    Since $z_1 = w^Ty_1 \sim \mathcal{N}(0, w^T\Sigma_1w)$ and $z_2 = w^Ty_2 \sim \mathcal{N}(0, w^T\Sigma_2w)$ we have:

    $$E(w) = \frac{1}{2}\mathbb{P}_{z_1 \sim \mathcal{N}(0, w^T\Sigma_1w)}(z_1 < -\mu_1^Tw) + \frac{1}{2}\mathbb{P}_{z_2 \sim \mathcal{N}(0, w^T\Sigma_2w)}(z_2 \ge - \mu_2^Tw) = $$
    $$ = \frac{1}{2}\mathbb{P}_{z'_1 \sim \mathcal{N}(0, 1)}(z'_1 < - \frac{\mu_1^Tw}{\sqrt{w^T \Sigma_1 w}}) + \frac{1}{2}\mathbb{P}_{z'_2 \sim \mathcal{N}(0, 1)}(z'_2 \ge -\frac{\mu_2^Tw}{ \sqrt{w^T \Sigma_2 w}}) = $$
    $$= \frac{1}{2}\mathbb{P}_{z'_1 \sim \mathcal{N}(0, 1)}(z'_1 > \frac{\mu_1^Tw}{\sqrt{w^T \Sigma_1 w}}) + \frac{1}{2}\mathbb{P}_{z'_2 \sim \mathcal{N}(0, 1)}(z'_2 \ge -\frac{\mu_2^Tw}{ \sqrt{w^T \Sigma_2 w}}) = $$
    $$= \frac{1}{2}Q(\frac{\mu_1^Tw}{\sqrt{w^T \Sigma_1 w}}) + \frac{1}{2}Q(-\frac{\mu_2^Tw}{ \sqrt{w^T \Sigma_2 w}})$$
\end{proof}
\subsection{Proof of Lemma \ref{lem:optimal}}
\begin{proof}

Note that, since the model for the data distribution is symmetric w.r.t. the transformation $\mu_1 \xleftrightarrow{} \mu_2$, the equality $\mu_1^Tw_{opt} = -\mu_2^Tw_{opt}$ has to hold for the optimal $w_{opt}$ from each of the parts (1)-(3) of the lemma. One therefore has:
$$E(w_{opt}) = \frac{1}{2}Q\left(\frac{\mu_1^Tw_{opt}}{\sigma\|w_{opt}\|}\right) + \frac{1}{2}Q\left(\frac{-\mu_2^Tw_{opt}}{\sigma\|w_{opt}\|}\right) = Q\left(\frac{\mu_1^Tw_{opt}}{\sigma\|w_{opt}\|}\right)  = Q\left(\frac{(\mu_1 - \mu_2)^Tw_{opt}}{2\sigma\|w_{opt}\|}\right)$$

Now let us proceed to proving each of the points (1)-(3) separately:

1. Note that for any $w$ satisfying the symmetry condition $\mu_1^Tw = -\mu_2^Tw$ it holds that 
$$E(w) = Q\left(\frac{(\mu_1 - \mu_2)^Tw}{2\sigma\|w\|}\right) \ge Q\left(\frac{(\mu_1 - \mu_2)^Tw_*}{2\sigma\|w_*\|}\right) = Q\left(\frac{\|\mu_1 - \mu_2\|_2}{2\sigma}\right) = Q\left(\sqrt{\frac{d(1-r)}{2\sigma^2}}\right) \text{ where } w_* = \mu_1 - \mu_2$$

2. Assuming that $\mu_1^Tw = -\mu_2^Tw$ and $w$ is a $1$ - bit vector we have 
$$E(w) = Q\left(\frac{(\mu_1 - \mu_2)^Tw}{2\sigma\|w\|}\right) \ge Q\left(\frac{(\mu_1 - \mu_2)^Tw_*}{2\sigma\|w_*\|}\right) = Q\left(\frac{\|\mu_1 - \mu_2\|_1}{2\sigma}\right) = Q\left(\sqrt{\frac{d(1-r)}{\pi\sigma^2}}\right)\text{ where } w_* = sign(\mu_1 - \mu_2)$$

3. If $w$ is a $k$ - sparse vector such that $\mu_1^Tw = -\mu_2^Tw$, then 
$$E(w) = Q\left(\frac{(\mu_1 - \mu_2)^Tw}{2\sigma\|w\|}\right) \ge Q\left(\frac{(\mu_1 - \mu_2)^Tw_*}{2\sigma\|w_*\|}\right)$$
where $w_*$ is obtained from taking the $k$ coordinates of $\mu_1 - \mu_2$ with the largest magnitude and zeroing out the rest. 

\end{proof}

\section{PROOFS OF THE MAIN RESULTS}

\subsection{Proof of Theorem \ref{thm: AO_gen}}
\begin{proof}
As discussed in Section \ref{subs: setting}, the proof reduces to analyzing the following optimization problem:
\begin{align*}
    \min_w \| (A+M) w -z  \|_2^2 + \lambda f(w)  
\end{align*}

We rewrite it as a min-max problem to enable invocation of CGMT: 
\begin{align*}
    \min_w \max_v   v^TAw + v^TMw - v^Tz - \frac{1}{4}\|v\|^2 + \lambda  f(w) 
\end{align*}
Applying CGMT yields the following AO:
\begin{align*}
    \min_w \max_v \|v\| g^Tw+ \|w\| h^Tv + v^T(Mw - z) - \frac{1}{4}\|v\|^2 + \lambda  f(w) 
\end{align*} 
Performing optimization over $v$ leads to:
\begin{align}\label{eq: ao_ge}
    \min_w \max_{\beta \geq 0} \beta g^Tw+ \beta \| h \|w\| + Mw - z \|- \frac{1}{4}\beta^2 + \lambda  f(w) 
\end{align}

Substituting for the norm of $\| h \|w\| + Mw - z \|$ we obtain:
\begin{align*}
    \min_w \max_{\beta \geq 0} \beta g^Tw+ \beta \sqrt{n \sigma^2 \|w\|^2 + \|Mw\|^2 + \| z\|^2 - 2z^T Mw }- \frac{1}{4}\beta^2 + \lambda  f(w) 
\end{align*}

Recall that by definition $M = \begin{pmatrix}
    \mathds{1}_{\frac{n}{2}} & 0 \\
    0 & \mathds{1}_{\frac{n}{2}}
\end{pmatrix} 
\begin{pmatrix}
    \mu_1^T \\
    \mu_1^T
\end{pmatrix}$. We then have $M^T M = \begin{pmatrix}
    \frac{n}{2} & 0 \\
    0 & \frac{n}{2}
\end{pmatrix}$ and therefore 
$$\|Mw\|^2 = \frac{n}{2}( (\mu_1^Tw)^2 + (\mu_2^Tw)^2)$$

Noting that $\| z\|^2 = n$  and using the square root trick we arrive to:
\begin{align*}
    \min_{w, \tau \geq 0} \max_{\beta \geq 0} \beta g^Tw+ \frac{\beta }{2\tau} + \frac{\beta\tau}{2} (n \sigma^2 \|w\|^2 + \frac{n}{2}(\mu_1^Tw)^2 + \frac{n}{2} (\mu_2^Tw)^2 + n - 2z^T Mw)- \frac{1}{4}\beta^2 + \lambda f(w)  
\end{align*}

Writing the Fenchel dual for each of $(\mu_1^Tw)^2$ and $(\mu_2^Tw)^2$ we get:
\begin{align*}
    \min_{w, \tau \geq 0} \max_{\beta \geq 0, \gamma_1, \gamma_2} \beta g^Tw+ \frac{\beta }{2\tau} + \frac{\beta\tau}{2} ( n \sigma^2 \|w\|^2 + \frac{n}{2} \gamma_1 (\mu_1^Tw)- \frac{n}{8} \gamma_1^2 + \frac{n}{2}\gamma_2 (\mu_2^Tw)- \frac{n}{8} \gamma_2^2 + n - 2z^T Mw)- \frac{1}{4}\beta^2 + \lambda  f(w) 
\end{align*}
Open $f(w)$ up as $\sum_{i=1}^d f_i(w_i)$. The objective above can be rewritten as: 
\begin{align*}
    &\min_{\tau \geq 0} \max_{\beta \geq 0, \gamma_1, \gamma_2} \frac{\beta }{2\tau} + \frac{\beta\tau}{2} ( - \frac{n}{8} \gamma_1^2 - \frac{n}{8} \gamma_2^2 + n )- \frac{1}{4}\beta^2 + \\
    & + \min_w \beta g^Tw +  \frac{\beta\tau n \sigma^2}{2} \|w\|^2 + \frac{\beta\tau n}{4} \gamma_1 (\mu_1^Tw) + \frac{\beta\tau n}{4}\gamma_2 (\mu_2^Tw) - \beta\tau z^T Mw + \lambda \sum_{i=1}^d f_i(w_i)
\end{align*}

We will recover the optimal $w$ now. Since the first part of the expression above is independent of $w$, we will put it aside for a time being and look only at the second one:
\begin{align}\label{eq:ao_separated}
    & \min_w \beta g^Tw +  \frac{\beta\tau n \sigma^2}{2} \|w\|^2 + \frac{n\beta\tau}{4} \gamma_1 (\mu_1^Tw) + \frac{n\beta\tau}{4}\gamma_2 (\mu_2^Tw) - \beta\tau z^T Mw + \lambda \sum_{i=1}^d f_i(w_i) = \nonumber\\
    & =  \sum_{i=1}^d \min_{w_i}  \frac{\beta\tau n \sigma^2}{2} w_i^2 + \beta (g +  \frac{n\tau}{4} \gamma_1 \mu_1 + \frac{n\tau}{4} \gamma_2  \mu_2 - \tau M^Tz )_i w_i + \lambda f_i(w_i) = \nonumber\\
    & = \sum_{i=1}^d \min_{w_i}  \frac{\beta\tau n \sigma^2}{2} (w_i + \frac{1}{n \tau \sigma^2} (g +  \frac{n\tau}{4} \gamma_1  \mu_1 + \frac{n\tau}{4} \gamma_2  \mu_2 - \tau M^Tz )_i)^2 + \lambda f_i(w_i)  - \frac{\beta}{2n \tau \sigma^2} (g +  \frac{n\tau}{4} \gamma_1  \mu_1 + \frac{n\tau}{4} \gamma_2  \mu_2 - \tau M^Tz )_i^2 
\end{align}
The expression above is equal to the following by definition of the Moreau envelope:
\begin{align*}
- \frac{\beta}{2n \tau \sigma^2} \sum_{i=1}^d (g +  \frac{n\tau}{4} \gamma_1  \mu_1 + \frac{n\tau}{4} \gamma_2  \mu_2 - \tau M^Tz )_i^2 + \lambda  \sum_{i=1}^d e_{ f_i} (\frac{1}{n \tau \sigma^2}(g +  \frac{n\tau}{4} \gamma_1  \mu_1 + \frac{n\tau}{4}  \mu_2 \gamma_2 - \tau M^Tz)_i ; \frac{\lambda}{\beta\tau n \sigma^2})
\end{align*}

We will simplify the objective further. Notice that also by definition $M^Tz = \frac{n}{2}(1-2c) (\mu_1 - \mu_2)$. This equality can be plugged in into the corresponding part of the objective:
\begin{align*}
     g +  \frac{n\tau}{4} \gamma_1  \mu_1 + \frac{n\tau}{4} \gamma_2  \mu_2 - \tau M^Tz  = g + (\frac{n\tau}{4} \gamma_1 - \frac{n\tau}{2} (1-2c) ) \mu_1 +  (\frac{n\tau}{4} \gamma_2 + \frac{n}{2}\tau (1-2c) ) \mu_2 
\end{align*}
Recall that $\mu_1$ and $\mu_2$ are i.i.d Gaussian by assumption. Therefore, we have the following equality asymptotically when $d \to \infty$, where $G \sim \mathcal{N}(0,1)$:
\begin{align*}
    &d \frac{1}{d }\sum_{i=1}^d e_{f_i} (\frac{1}{n \tau \sigma^2}(g +  \frac{n\tau}{4} \gamma_1  \mu_1 + \frac{n\tau}{4}  \mu_2 \gamma_2 - \tau M^Tz)_i;\frac{\lambda}{\beta\tau n \sigma^2}) \rightarrow \\
    &d \mathbb{E } e_{f_i} (\frac{1}{n \tau \sigma^2}\sqrt{(\frac{n\tau}{4} \gamma_1 -\frac{n\tau}{2} (1-2c))^2 +  (\frac{n\tau}{4} \gamma_2 + \frac{n\tau}{2} (1-2c) )^2 + 2r (\frac{n\tau}{4} \gamma_1 - \frac{n\tau}{2} (1-2c)) (\frac{n\tau}{4} \gamma_2 +\frac{n\tau}{2} (1-2c) ) + \sigma^2} G;\frac{\lambda}{\beta\tau n \sigma^2} )
\end{align*}

Note that we can take  $\gamma_1 = \gamma = - \gamma_2$ because our model for the data is symmetric w.r.t to the substitution $\mu_1 \xleftrightarrow{} -\mu_2$. Thus, the expression above simplifies into:
\begin{align*} 
d \mathbb{E } e_{ f_i} (\sqrt{\frac{1-r}{2\sigma^4}(\frac{\gamma}{2} -1+2c)^2  + \frac{1}{n^2 \tau^2 \sigma^2}} G; \frac{\lambda}{\beta\tau n \sigma^2})
\end{align*}
Note also that
\begin{align*}
     &\sum_{i=1}^d  (g + (\frac{n\tau}{4} \gamma -\frac{n\tau}{2} (1-2c) ) (\mu_1 - \mu_2) )_i^2 = \| g\|^2 +  (\frac{n\tau}{4} \gamma - \frac{n\tau}{2} (1-2c) )^2 \| \mu_1 - \mu_2\|^2 = d \sigma^2 + 2 d (1-r)  (\frac{n\tau}{4} \gamma - \frac{n\tau}{2} (1-2c) )^2
\end{align*}
Putting all derivations above together, we are left with the following optimization problem:
\begin{align*}
    &\min_{\tau \geq 0} \max_{\beta \geq 0, \gamma} \frac{\beta }{2\tau} + \frac{\beta\tau}{2} ( - \frac{n}{4} \gamma^2 + n)- \frac{1}{4}\beta^2 - \frac{d \beta}{2n \tau} - \frac{nd (1-r) \beta \tau}{4\sigma^2} (\frac{\gamma}{2} - 1 + 2c)^2\\
    & + d \lambda \mathbb{E } e_{ f_i} (\sqrt{\frac{1-r}{2\sigma^4}(\frac{\gamma}{2} -1+2c)^2  + \frac{1}{n^2 \tau^2 \sigma^2}} G; \frac{\lambda}{\beta\tau n \sigma^2})
\end{align*}
Denoting $\Xi(\gamma, \tau) := \sqrt{\frac{1-r}{2\sigma^4}(\frac{\gamma}{2} -1+2c)^2  + \frac{1}{n^2 \tau^2 \sigma^2}}$, we arrive to the desired result:
\begin{align*}
    &\min_{\tau \geq 0} \max_{\beta \geq 0, \gamma} \frac{\beta }{2\tau} (1 - \frac{d}{n}) + \frac{\beta\tau}{2} ( - \frac{n}{4} \gamma^2 - \frac{nd (1-r)}{2\sigma^2} (\frac{\gamma}{2} - 1 + 2c)^2  + n)- \frac{1}{4}\beta^2 \\
    & + d \lambda \mathbb{E } e_{ f_i} (\Xi(\gamma, \tau) G; \frac{\lambda}{\beta\tau n \sigma^2})
\end{align*}
\end{proof}
\subsection{Proof of Theorem \ref{thm: ridge}}
\begin{proof}
Apply the same reasoning as in the proof of Theorem \ref{thm: AO_gen} up to the equation (\ref{eq:ao_separated}). We have: 
\begin{align*}
\sum_{i=1}^d \min_{w_i}  \frac{\beta\tau n \sigma^2}{2} (w_i + \frac{1}{n \tau \sigma^2} (g +  \frac{n\tau}{4} \gamma_1  \mu_1 + \frac{n\tau}{4} \gamma_2  \mu_2 - \tau M^Tz )_i)^2 + \lambda w_i^2  - \frac{\beta}{2n \tau \sigma^2} (g +  \frac{n\tau}{4} \gamma_1  \mu_1 + \frac{n\tau}{4} \gamma_2  \mu_2 - \tau M^Tz )_i^2 
\end{align*}

Taking the derivative by $w$ of the expression above and setting it to $0$ we get:

$\beta\tau n \sigma^2 (w + \frac{1}{n \tau \sigma^2} (g +  \frac{n\tau}{4} \gamma_1  \mu_1 + \frac{n\tau}{4} \gamma_2  \mu_2 - \tau M^Tz )) + 2\lambda w = 0$

Since $M^Tz = \frac{n}{2}(1-2c) (\mu_1 - \mu_2)$, we arrive at:
$$w = -\frac{\beta\tau n \sigma^2 (g +  \frac{n\tau}{4} \gamma_1  \mu_1 + \frac{n\tau}{4} \gamma_2  \mu_2 - \frac{n\tau}{2}(1-2c) (\mu_1 - \mu_2) )}{n \tau \sigma^2(\beta\tau n \sigma^2 + 2\lambda)}$$

We conclude from the identity above that $w$ belongs in the span of $g, \mu_1$ and $\mu_2$. Moreover, since this identity is invariant under the transformation $(\gamma_1, \gamma_2, \mu_1, \mu_2) \to (-\gamma_2, \gamma_1, -\mu_2, \mu_1)$, we also conclude that $\gamma_1 = - \gamma_2$. Put together, this implies that $w$ can be written as $w = \alpha g + \gamma(\mu_1 - \mu_2)$. Now return to the equation (\ref{eq: ao_ge}) derived the proof of Theorem \ref{thm: AO_gen} and specify it to the case $f = \ell_2$: 
\begin{align*}
    \min_w \max_{\beta \geq 0} \beta g^Tw+ \beta \| h \|w\| + Mw - z \|- \frac{1}{4}\beta^2 + \lambda  \|w\|^2 
\end{align*}
Performing optimization over $\beta$ we obtain:
\begin{align}\label{eq: ao_ridge}
    \min_w (g^Tw+ \| h \|w\| + Mw - z \|)_{\ge 0}^2 + \lambda \|w\|_2^2
\end{align}
Denote:
\begin{align*}
\Omega(\alpha, \gamma)  &=   \alpha ^ 2 \sigma^2 d + 2d(1-r)\gamma ^ 2 \\
\Theta(\alpha, \gamma)  &=   n(1-c)(d(1-r)\gamma - 1) ^ 2 + nc(d(1-r)\gamma + 1) ^ 2   
\end{align*}
It is straightforward to open up each term from the objective above in terms of $\alpha$ and $\gamma$:
\begin{align*}
& g^Tw = g^T(\alpha g + \gamma(\mu_1 - \mu_2)) = \alpha\sigma^2d \\
& \|w\|^2 = \|\alpha g + \gamma(\mu_1 - \mu_2)\| ^ 2 = \alpha ^ 2 \sigma^2 d + 2d(1-r)\gamma ^ 2 = \Omega(\alpha, \gamma)\\
&  \| h \|w\| + Mw - z \| = \sqrt{n\sigma^2\|w\|^2 + \|Mw - z \|^2} = \\
& = \sqrt{n\sigma^2\|w\|^2 + \frac{n(1-c)}{2}((\mu_1^Tw-1)^2 + (\mu_2^Tw+1)^2) + \frac{nc}{2}((\mu_1^Tw+1)^2 + (\mu_2^Tw-1)^2)} = \\
& = \sqrt{n \sigma ^ 2 \Omega(\alpha, \gamma) + \Theta(\alpha, \gamma)}
\end{align*}
Plugging these into equation (\ref{eq: ao_ridge}), we get the desired objective: 
$$ \min_{\alpha, \gamma} (\alpha\sigma^2d + \sqrt{n \sigma ^ 2 \Omega(\alpha, \gamma) + \Theta(\alpha, \gamma)})_{\ge 0} ^ 2 + \lambda \Omega(\alpha, \gamma)$$
Using Lemma \ref{lem:gen_error} we also derive:
$$E(w) = \frac{1}{2}Q\left(\frac{\mu_1^Tw}{\sigma\|w\|}\right) + \frac{1}{2}Q\left(\frac{-\mu_2^Tw}{\sigma\|w\|}\right) = Q\left(\frac{\mu_1^Tw}{\sigma\|w\|}\right) = Q\left(\frac{\mu_1^Tw}{\sqrt{\sigma^2 \Omega(\alpha, \gamma)}}\right) = Q\left(\frac{\gamma d(1 - r)}{\sqrt{\alpha ^ 2\sigma ^ 4 d + 2\sigma ^ 2  \gamma ^ 2 d(1 - r)}}\right)$$
\end{proof}
\subsection{Proof of Theorem \ref{thm: AO_l1}}
\begin{proof}
Starting from the equation (\ref{eq:ao_separated}) derived in the proof of Theorem \ref{thm: AO_gen} applied to $f = \ell_1$:
\begin{align*}
    \min_{w, \tau \geq 0} \max_{\beta \geq 0} \beta g^Tw+ \frac{\beta }{2\tau} + \frac{\beta\tau}{2} ( n \sigma^2 \|w\|^2 + \frac{n}{2}(\mu_1^Tw)^2 + \frac{n}{2} (\mu_2^Tw)^2 + n - 2z^T Mw)- \frac{1}{4}\beta^2 + \lambda \|w\|_1
\end{align*}

Using Fenchel duality and introducing $\gamma_1, \gamma_2$ yields the following expression:
\begin{align*}
     \min_{w, \tau \geq 0} \max_{\beta \geq 0, \gamma_1, \gamma_2} \beta g^Tw+ \frac{\beta }{2\tau} + \frac{\beta\tau}{2} ( n \sigma^2 \|w\|^2 + \frac{n}{2} \gamma_1 (\mu_1^Tw)- \frac{n}{8} \gamma_1^2 + \frac{n}{2}\gamma_2 (\mu_2^Tw)- \frac{n}{8} \gamma_2^2 + n - 2z^T Mw)- \frac{1}{4}\beta^2 + \lambda \|w\|_1
\end{align*}
Due to the symmetry to the transformation $(\gamma_1, \gamma_2, \mu_1, \mu_2) \to (-\gamma_2, \gamma_1, -\mu_2, \mu_1)$, we have $\gamma_1 = -\gamma_2 =: \gamma $ leading to the following optimization problem:
\begin{align*}
    \min_{w, \tau \geq 0} \max_{\beta \geq 0, \gamma} \beta g^Tw+ \frac{\beta }{2\tau} + \frac{\beta\tau}{2} ( n \sigma^2 \|w\|^2
    + n \gamma (\mu_1^Tw)- \frac{n}{4} \gamma^2 + n - 2z^T M)- \frac{1}{4}\beta^2 + \lambda  \| w\|_1 
\end{align*}
Solving for $w$ we get:
\begin{align*}
    w_{opt} &=
    \begin{cases}
        \frac{X}{\sigma^2} - \lambda \frac{1}{n\beta\tau \sigma^2} & X  > \lambda \frac{1}{n\beta\tau} \\ 
        \frac{X}{\sigma^2} + \lambda \frac{1}{n\beta\tau\sigma^2} & X < -\lambda \frac{1}{n\beta\tau} \\
        0 & o.w
    \end{cases}
\end{align*}
Where
\begin{align*} 
X &= -\frac{1}{n\tau}g_i + (\frac{1-2c}{2}  - \frac{\gamma}{4})(\mu_{1,i} - \mu_{2,i}) \sim \mathcal{N}(0, \tilde{\sigma}^2) \\
\tilde{\sigma}^2 &= \frac{\sigma^2}{n^2\tau^2} + 2(\frac{\gamma}{4} - \frac{1-2c}{2})^2(1-r) 
\end{align*}

The optimization would then be
\begin{align*}
    \min_{\tau \geq 0} \max_{\beta \geq 0, \gamma} \beta g^Tw_{opt}+ \frac{\beta }{2\tau} + \frac{\beta\tau}{2} ( n \sigma^2 \|w_{opt}\|^2
    + n \gamma (\mu_1^Tw_{opt})- \frac{n}{4} \gamma^2 + n - 2z^T Mw_{opt})- \frac{1}{4}\beta^2 + \lambda  \| w_{opt}\|_1 
\end{align*}

Now we will plug in $w_{opt}$ and calculate each term. The first term concentrates to
\begin{align*}
    g ^ T w_{opt} = -\frac{2d}{n\tau} Q(\frac{\lambda}{n \beta \tau \tilde{\sigma}})
\end{align*}
Similarly, we have for $\mu_1 ^ Tw_{opt}$ and $z^TMw_{opt}$:
\begin{align*}
    \mu_1 ^ Tw_{opt} = d \mathbb{E} \mu_{1,i}w_i = \frac{d}{\sigma^2} \mathbb{E} \mu_{1,i}X = \frac{2d}{\sigma^2}Q(\frac{\lambda}{n \beta \tau \tilde{\sigma}}) (\frac{1-2c}{2} - \frac{\gamma}{4})(1-r)
\end{align*}
\begin{align*}
    z^TMw_{opt} = \frac{n(1-2c)}{2}(\mu_1 - \mu_2)^Tw_{opt} = \frac{2dn(1-2c)}{\sigma^2}Q(\frac{\lambda}{n \beta \tau \tilde{\sigma}}) (\frac{1-2c}{2} - \frac{\gamma}{4})(1-r)
\end{align*}
For $\|w_{opt}\|_1$ and $\|w_{opt}\| ^ 2$ we get:
\begin{align*}
    &\|w_{opt}\|_1 = d (\mathbb{E} (\frac{X}{\sigma^2} -  \frac{\lambda}{n\beta\tau \sigma^2})\mathds{1}_{X  > \frac{\lambda }{n\beta\tau}} + \mathbb{E} (-\frac{X}{\sigma^2} -  \frac{\lambda}{n\beta\tau \sigma^2})\mathds{1}_{X  < - \frac{\lambda}{n\beta\tau}}) = \\
    & = \frac{2d\tilde{\sigma}}{\sigma^2\sqrt{2\pi}} e^{-\frac{\lambda^2}{2(n\beta\tau\tilde{\sigma})^2}} -  \frac{2d\lambda}{n\beta\tau \sigma^2} Q(\frac{\lambda}{n\beta\tau\tilde{\sigma}})
\end{align*}
Denoting  $s := \frac{\tilde{\sigma}n\beta\tau}{\lambda}$
\begin{align*}
    \|w_{opt}\| ^ 2 = d\mathbb{E}(\frac{X}{\sigma^2} - \frac{\lambda}{n \beta \tau\sigma^2}) ^ 2 \mathds{1}_{|X| \ge \frac{\lambda}{n \beta \tau}} = d(\frac{\lambda}{n\beta\tau\sigma^2})^2[2(s^2+1)Q(\frac{1}{s}) - \frac{2s}{\sqrt{2\pi}}e^{-\frac{1}{2s^2}}] 
\end{align*}
Putting the results together, we arrive at the following objective:
\begin{align*}
    \min_{\tau \geq 0} \max_{\beta \geq 0, \gamma} -\frac{2d\beta}{n\tau} Q(\frac{\lambda}{n \beta \tau \tilde{\sigma}}) + \frac{\beta}{2\tau} + \frac{\beta\tau n}{2}(d\sigma^2(\frac{\lambda}{n\beta\tau\sigma^2})^2[  2(s^2+1)Q(\frac{1}{s}) - \frac{2s}{\sqrt{2\pi}}e^{-\frac{1}{2s^2}}] + \frac{2\gamma d}{\sigma^2}Q(\frac{\lambda}{n \beta \tau \tilde{\sigma}}) (\frac{1-2c}{2} - \frac{\gamma}{4})(1-r)) + \\
    + \frac{\beta\tau n}{2} (-\frac{\gamma^2}{4}  + 1 -  \frac{4d(1-2c)}{\sigma^2}Q(\frac{\lambda}{n \beta \tau \tilde{\sigma}}) (\frac{1-2c}{2} - \frac{\gamma}{4})(1-r))- \frac{1}{4}\beta^2 +
   \frac{2d\lambda\tilde{\sigma}}{\sigma^2\sqrt{2\pi}} e^{-\frac{\lambda^2}{2(n\beta\tau\tilde{\sigma})^2}} -  \frac{2d\lambda^2}{n\beta\tau \sigma^2} Q(\frac{\lambda}{n\beta\tau\tilde{\sigma}})
\end{align*}

To calculate the sparsity rate, we analyze the probability of $w_{opt,i}$ attaining non-zero values which is equivalent to the occurrence of the event  $\{ |X| >  \frac{\lambda}{n\beta\tau} \}$:
\begin{align*}
    \mathbb{P}(w_i \ne 0) = \mathbb{P}\left(|X| >  \frac{\lambda}{n\beta\tau}\right) 
\end{align*}
A straightforward calculation of integrals yields $\mathbb{P}(|X| > \frac{\lambda}{n \beta \tau}) = 2 Q(\frac{\lambda}{n \beta \tau \tilde{\sigma}})$ and therefore $\mathbb{P}(w_i \ne 0) = 2 Q(\frac{\lambda}{n \beta \tau \tilde{\sigma}})$. Subsequently, the solution is $k$-sparse, where $k = \lfloor 2 d Q(\frac{\lambda}{n \beta \tau \tilde{\sigma}}) \rfloor $.
\end{proof}

\subsection{Proof of Theorem \ref{thm: AO_linf}}
\begin{proof}
Starting from equation (\ref{eq:ao_separated}) again, we introduce the $\ell_{\infty}$ regularization term as a constraint using a new variable $\delta$:
\begin{align*}
    \min_{\|w\|_{\infty} \leq \frac{\delta}{\lambda}, \delta, \tau \geq 0} \max_{\beta \geq 0} \beta g^Tw+ \frac{\beta }{2\tau} + \frac{\beta\tau}{2} ( n \sigma^2 \|w\|^2 + \frac{n}{2}(\mu_1^Tw)^2 + \frac{n}{2} (\mu_2^Tw)^2 + n - 2z^T Mw)- \frac{1}{4}\beta^2 + \delta
\end{align*}

Using Fenchel duality and introducing $\gamma_1, \gamma_2$ yields the following expression:
\begin{align*}
     \min_{\|w\|_{\infty} \leq \frac{\delta}{\lambda}, \delta, \tau \geq 0} \max_{\beta \geq 0, \gamma_1, \gamma_2} \beta g^Tw+ \frac{\beta }{2\tau} + \frac{\beta\tau}{2} ( n \sigma^2 \|w\|^2 + \frac{n}{2} \gamma_1 (\mu_1^Tw)- \frac{n}{8} \gamma_1^2 + \frac{n}{2}\gamma_2 (\mu_2^Tw)- \frac{n}{8} \gamma_2^2 + n - 2z^T Mw)- \frac{1}{4}\beta^2 + \delta
\end{align*}

Due to the same symmetry as in the proof of the previous theorem, it holds that $\gamma_1 = -\gamma_2 =: \gamma$ and the objective turns into
\begin{align*}
     \min_{\|w\|_{\infty} \leq \frac{\delta}{\lambda}, \delta, \tau \geq 0} \max_{\beta \geq 0, \gamma} \beta g^Tw+ \frac{\beta }{2\tau} + \frac{\beta\tau}{2} ( n \sigma^2 \|w\|^2 + \frac{n}{2} \gamma(\mu_1^Tw)- \frac{n}{4} \gamma^2 - \frac{n}{2}\gamma (\mu_2^Tw) + n - 2z^T Mw)- \frac{1}{4}\beta^2 + \delta
\end{align*}

Now we can take derivative w.r.t $w$ and find the optimal solution, $w_{opt}$
\begin{align*} (w_{opt})_i =
    \begin{cases}
        \frac{\delta}{\lambda} & \frac{1}{{n\tau \sigma^2}} (\tau M^T z -  g  - \frac{n\tau\gamma}{4} \mu_1 +  \frac{n\tau\gamma}{4} \mu_2)_i > \frac{\delta}{\lambda} \\
        \frac{1}{{n\tau \sigma^2}} (\tau M^T z -  g  - \frac{n\tau\gamma}{4} \mu_1 +  \frac{n\tau\gamma}{4} \mu_2)_i & -\frac{\delta}{\lambda} \leq \frac{1}{{n\tau \sigma^2}} (\tau M^T z -  g  - \frac{n\tau\gamma}{4} \mu_1 +  \frac{n\tau\gamma}{4} \mu_2)_i \leq \frac{\delta}{\lambda} \\
        -\frac{\delta}{\lambda} & \frac{1}{{n\tau \sigma^2}} (\tau M^T z -  g  - \frac{n\tau\gamma}{4} \mu_1 +  \frac{n\tau\gamma}{4} \mu_2)_i < -\frac{\delta}{\lambda}
    \end{cases}
\end{align*}
We note that $M^T z = \frac{n}{2}(1-2c) (\mu_1 - \mu_2)$ and thus
\begin{align*}
     g +  \frac{n\tau}{4} \gamma  \mu_1 - \frac{n\tau}{4} \gamma  \mu_2 - \tau M^Tz  = g + (\frac{n\tau}{4} \gamma- \frac{n\tau}{2} (1-2c) ) \mu_1 +  (-\frac{n\tau}{4} \gamma + \frac{n}{2}\tau (1-2c) ) \mu_2 
\end{align*}
Then plugging in the previous expression,
\begin{align*} (w_{opt})_i =
    \begin{cases}
        \frac{\delta}{\lambda} & \frac{1}{{ 2\sigma^2}} (-(\frac{\gamma}{2}- 1 + 2c ) (\mu_1 - \mu_2) - \frac{2}{n\tau} g )_i  > \frac{\delta}{\lambda} \\
        \frac{1}{{ 2\sigma^2}} (-(\frac{\gamma}{2}- 1 + 2c ) (\mu_1 - \mu_2) - \frac{2}{n\tau} g )_i  & -\frac{\delta}{\lambda} \leq \frac{1}{{ 2\sigma^2}} (-(\frac{\gamma}{2}- 1 + 2c ) (\mu_1 - \mu_2) - \frac{2}{n\tau} g )_i \leq \frac{\delta}{\lambda} \\
        -\frac{\delta}{\lambda} & \frac{1}{{ 2\sigma^2}} (-(\frac{\gamma}{2}- 1 + 2c ) (\mu_1 - \mu_2) - \frac{2}{n\tau} g )_i  < -  \frac{\delta}{\lambda}
    \end{cases}
\end{align*}
Let $\Xi(\gamma, \tau) := \sqrt{\frac{1-r}{2\sigma^4}(\frac{\gamma}{2} -1+2c)^2  + \frac{1}{n^2 \tau^2 \sigma^2}}$. Using the assumptions on $\mu_1, \mu_2$ we model the randomness coming from $\mu_1, \mu_2, g$ by a single $G_i \sim \mathcal{N}(0,1)$:
\begin{align*} (w_{opt})_i =
    \begin{cases}
        \frac{\delta}{\lambda} & \Xi(\gamma, \tau) G_i  > \frac{\delta}{\lambda} \\
        \Xi(\gamma, \tau) G_i  & -\delta \leq \Xi(\gamma, \tau) G_i \leq \frac{\delta}{\lambda} \\
        -\frac{\delta}{\lambda} & \Xi(\gamma, \tau) G_i  < -  \frac{\delta}{\lambda}
    \end{cases}
\end{align*}

After finding $w_{opt}$ we plug its value into the original optimization
\begin{align*}
     &\min_{\tau \geq 0} \max_{\beta \geq 0, \gamma} \frac{\beta }{2\tau} + \frac{\beta\tau}{2} ( - \frac{n}{4} \gamma^2 + n ) - \frac{1}{4}\beta^2 + \delta \\
     & + \frac{n \sigma^2\beta\tau}{2} \|w_{opt}\|^2 + (\beta g  + \frac{\beta\tau n}{4} \gamma \mu_1 - \frac{\beta\tau n}{4} \mu_2 - \beta\tau M^Tz)^T w_{opt} 
\end{align*}

It remains to express $\|w_{opt}\|^2$ and $(\beta g  + \frac{\beta\tau n}{4} \gamma \mu_1 - \frac{\beta\tau n}{4} \mu_2 - \beta\tau M^Tz)^T w_{opt} $ in terms of the scalars. For $\|w_{opt}\|^2$ we have
\begin{align*}
    & \|w_{opt}\|^2 = \frac{d}{\sqrt{2\pi}} \int_{\lambda \Xi |G| \geq \delta} \frac{\delta^2}{\lambda^2} exp(- \frac{G^2}{2}) dG + \frac{d \Xi^2}{\sqrt{2\pi}} \int_{\lambda\Xi |G| \leq \delta} G^2 exp(- \frac{G^2}{2}) dG = \\
    & \frac{2d\delta^2}{\lambda^2} Q(\frac{\delta}{\lambda\Xi}) - \frac{2d \delta \Xi}{\lambda\sqrt{2\pi}} exp(- \frac{\delta^2}{2\lambda^2\Xi^2}) + d \Xi^2 (1-2Q(\frac{\delta}{\lambda \Xi}))
\end{align*}

For inner product we observe the following concentration phenomenon:
\begin{align*}
    &( g  + \frac{n\tau}{4}  \gamma_1 \mu_1 + \frac{n\tau}{4}\gamma_2 \mu_2 - \tau M^Tz)^T w_{opt} = (g + (\frac{n\tau}{4} \gamma_1 - \frac{n\tau}{2} (1-2c) ) \mu_1 +  (\frac{n\tau}{4} \gamma_2 + \frac{n}{2}\tau (1-2c) ) \mu_2 )^T w_{opt} 
    \end{align*}
 \[  = - d n \sigma^2 \tau \Xi \mathbb{E} G w_{opt,i}\]

Thus we have:
\begin{align*}
     \mathbb{E} G w_{opt,i} = \frac{ \delta}{\lambda \sqrt{2\pi}} \int_{\lambda\Xi G \geq \delta} G exp(- \frac{G^2}{2}) dG - \frac{\delta}{\lambda\sqrt{2\pi}} \int_{\lambda\Xi G \leq -\delta} G exp(- \frac{G^2}{2}) dG  + \frac{\Xi}{\sqrt{2\pi}} \int_{\lambda\Xi |G| \leq \delta} G^2 exp(- \frac{G^2}{2}) dG =
\end{align*}
$$=\Xi (1-2Q(\frac{\delta}{\lambda \Xi}))$$
Therefore:
\begin{align*}
    & \frac{n \sigma^2\beta\tau}{2} \|w_{opt}\|^2 + (\beta g  + \frac{\beta\tau}{2} \frac{n}{2} \gamma_1 \mu_1 + \frac{\beta\tau}{2} \frac{n}{2}\gamma_2 \mu_2 - \beta\tau M^Tz)^T w_{opt} = \\
    & - \frac{n d \sigma^2\beta\tau \delta \Xi}{\lambda\sqrt{2\pi}} exp(- \frac{\delta^2}{2\lambda^2\Xi^2}) + n d \sigma^2\beta\tau ( \Xi^2 + \frac{\delta^2}{\lambda^2}) Q (\frac{\delta}{\lambda \Xi}) - \frac{n d \sigma^2\beta\tau \Xi^2}{2}
\end{align*}
The optimization problem turns into:
\begin{align*}
     &\min_{\tau, \delta \geq 0} \max_{\beta \geq 0, \gamma} \frac{\beta }{2\tau} + \frac{\beta\tau}{2} ( - \frac{n}{4} \gamma^2 + n )- \frac{1}{4}\beta^2 + \delta \\
     & - \frac{n d \sigma^2\beta\tau \delta \Xi}{\lambda\sqrt{2\pi}} exp(- \frac{\delta^2}{2\lambda^2\Xi^2}) + n d \sigma^2\beta\tau ( \Xi^2 + \frac{\delta^2}{\lambda^2}) Q (\frac{\delta}{\lambda \Xi}) - \frac{n d \sigma^2\beta\tau \Xi^2}{2} 
\end{align*}

Performing the optimization over $\beta$ we get the final expression:
\begin{align*}
     &\min_{\tau, \delta \geq 0} \max_{\gamma} \left( \frac{1 }{2\tau} + \frac{\tau}{2} ( - \frac{n}{4} \gamma^2 + n )  - \frac{n d \sigma^2\tau \delta \Xi}{\lambda\sqrt{2\pi}} exp(- \frac{\delta^2}{2\lambda^2\Xi^2}) + n d \sigma^2\tau ( \Xi^2 + \frac{\delta^2}{\lambda^2}) Q (\frac{\delta}{\lambda \Xi}) - \frac{n d \sigma^2\tau \Xi^2}{2}\right)^2_{\geq 0} + \delta 
\end{align*}

Considering the expression for $w_{opt}$, it can also be seen that
\begin{align*}
    d \mathbb{P} ( |w_{opt,i}| =\frac{\delta}{\lambda}) =  d \mathbb{P} ( | \Xi G | \geq \frac{\delta}{\lambda}) = \frac{d}{\sqrt{2\pi}} \int_{\lambda \Xi |G| \geq \delta} exp(- \frac{G^2}{2}) dG = 2d Q(\frac{\delta}{\lambda\Xi}) 
\end{align*}
\end{proof}
\subsection{Calculating the generalization error}
\begin{proof}
As an attentive reader might have noticed, we omitted derivations for the expressions for the generalization error from the proofs of Theorems \ref{thm: AO_gen}, \ref{thm: AO_l1}, \ref{thm: AO_linf}. The purpose of this subsection is filling in this gap.

Applying Lemma \ref{lem:gen_error} and using that $\mu_1^Tw = - \mu_2^Tw$ holds for the optimal $w$ from each of Theorems \ref{thm: AO_gen}, \ref{thm: AO_l1}, \ref{thm: AO_linf}, we have $E(w) = Q(\frac{\mu_1^Tw}{\sigma \|w\|})$. Thus, our goal is to determine the values of $\mu_1^Tw$ and $\|w\|$. The former is immediate, as $\gamma = 2\mu_1^Tw$. The latter is less trivial but still straightforward; by definition, 
$$n \sigma ^ 2\|w\| ^ 2 + \|Mw-z\| ^ 2 = \frac{1}{\tau ^ 2}$$ and $$\|Mw-z\| ^ 2 = \frac{n}{2}(1-c)(\mu_1 ^ T w - 1) ^ 2 + \frac{n}{2}(1-c)(\mu_2 ^ T w + 1) ^ 2 + \frac{nc}{2}(\mu_1 ^ T w + 1) ^ 2 + \frac{nc}{2}(\mu_2 ^ T w - 1) ^ 2 = $$
$$ = n(1-c)(\frac{\gamma}{2} - 1) ^ 2 + nc (\frac{\gamma}{2} + 1) ^ 2 $$

We derive
$$\|w\| ^ 2 = \frac{\frac{1}{\tau ^ 2} - n(1-c)(\frac{\gamma}{2} - 1) ^ 2 - nc (\frac{\gamma}{2} + 1) ^ 2}{n\sigma^2} $$

Putting the equalities above together yields
$$E(w) = Q(\frac{\gamma}{2\sigma \|w\|}) = Q(\frac{\gamma \sigma \sqrt{n}}{2\sigma \sqrt{\frac{1}{\tau ^ 2} - n(1-c)(\frac{\gamma}{2} - 1) ^ 2 - nc (\frac{\gamma}{2} + 1)^2}}) = Q(\frac{\gamma }{2\sqrt{\frac{1}{n\tau ^ 2} - (1-c)(\frac{\gamma}{2} - 1) ^ 2 - c (\frac{\gamma}{2} + 1)^2}})$$
\end{proof}
\section{DERIVATIONS BEHIND THE REMARKS}

Recall from Section \ref{subs: setting} that the problem is equivalent to analyzing the following:
\begin{align}\label{eq: po}
    \min_w \| (A+M) w -z  \|_2^2 + \lambda f(w)  
\end{align}
To gain more insight, we would like to apply a unitary transformation $U$, such that $$UM = \begin{pmatrix} \sqrt{\frac{n}{2}}\mu_1^T \\
\sqrt{\frac{n}{2}}\mu_2^T \\
0 \dots 0 \\
\dots \\
0 \dots 0
\end{pmatrix} \text{ and } Uz= \begin{pmatrix}
   \sqrt{\frac{n}{2}}(1-2c) \\
    -\sqrt{\frac{n}{2}}(1-2c) \\
    2 \sqrt{c(1-c)n} \\
    0 \\
    \dots \\
    0 \\
\end{pmatrix}$$

To explain why it exists, note that the desired matrix equality $UM = \begin{pmatrix} \sqrt{\frac{n}{2}}\mu_1^T \\
\sqrt{\frac{n}{2}}\mu_2^T \\
0 \dots 0 \\
\dots \\
0 \dots 0
\end{pmatrix}$ is guaranteed by the following two vector equalities: 
$$U\begin{pmatrix} 0_{\frac{n}{2}} \\
1_{\frac{n}{2}}
\end{pmatrix} = \begin{pmatrix} \sqrt{\frac{n}{2}} \\
0 \\
\dots\\ 
0 
\end{pmatrix} \text{ and }U\begin{pmatrix} 1_{\frac{n}{2}} \\
0_{\frac{n}{2}}
\end{pmatrix} = \begin{pmatrix} 0 \\
\sqrt{\frac{n}{2}} \\
0 \\
\dots\\ 
0 
\end{pmatrix}$$

Due to the defining property of a unitary operator, such a unitary $U$ exists if and only if it would preserve all pairwise dot products between the vectors it is defined on. It is easy to see that all the coordinates of the vectors in the image of $U$ are chosen so that this would indeed be true.

Apply $U$ to the term under the norm in the equation (\ref{eq: po}). We have: 

\begin{align}\label{eq: a}
    \min_w \left \| UAw + \begin{pmatrix} \sqrt{\frac{n}{2}}\mu_1^T \\
\sqrt{\frac{n}{2}}\mu_2^T \\
0 \dots 0 \\
\dots \\
0 \dots 0
\end{pmatrix}w -  \begin{pmatrix}
   \sqrt{\frac{n}{2}}(1-2c) \\
    -\sqrt{\frac{n}{2}}(1-2c) \\
    2 \sqrt{c(1-c)n} \\
    0 \\
    \dots \\
    0 \\
\end{pmatrix} \right \|_2^2 + \lambda f(w)  
\end{align}

Since $UA$ has the same distribution as $A$, it can be written as $UA = \begin{pmatrix} a_1 \\ a_2 \\a_3\\ \tilde{A}\end{pmatrix}$, where $a_i \sim \mathcal{N}(0, \sigma^2 I)$ for each $i = 1, 2, 3 $ and $\tilde{A} \in \mathbb{R}^{(n-3) \times d}$ has its entries $\tilde{A}_{ij} \sim \mathcal{N}(0, \sigma^2)$. Also denoting $\tilde{\mu_1} = \mu_1 + \sqrt{\frac{2}{n}}a_1$, $\tilde{\mu_2} = \mu_2 + \sqrt{\frac{2}{n}}a_2$ and $a=a_3$, we transform (\ref{eq: a}) into the following:

\begin{align*}
    \min_w \|\Tilde{A}w\|_2^2 + \frac{n}{2} (\Tilde{\mu_1}^T w - (1-2c))^2 + \frac{n}{2} (\Tilde{\mu_2}^T w + (1-2c))^2 + n (\frac{1}{\sqrt{n}}a^T w - 2 \sqrt{c(1-c)})^2 + \lambda f(w)
\end{align*}

Applying CGMT to the objective above in the same way as in the proof of Theorem \ref{thm: AO_gen} yields:
\begin{align}\label{eq: ao_detailed}
    \min_w (g^T w + \sqrt{n} \sigma \|w\|)_{\ge 0}^2 + \frac{n}{2} (\Tilde{\mu_1}^T w - (1-2c))^2 + \frac{n}{2} (\Tilde{\mu_2}^T w + (1-2c))^2 + n (\frac{1}{\sqrt{n}}a^T w - 2 \sqrt{c(1-c)})^2 + \lambda f(w)
\end{align}

\subsection{Derivations behind Remark \ref{rem: l2closed}}

Specifying equation (\ref{eq: ao_detailed}) to the case of $f = \ell_2$ we get:
\begin{align}\label{eq: ao_detailed_ridge}
    \min_w (g^T w + \sqrt{n} \sigma \|w\|)_{\ge 0}^2 + \frac{n}{2} (\Tilde{\mu_1}^T w - (1-2c))^2 + \frac{n}{2} (\Tilde{\mu_2}^T w + (1-2c))^2 + n (\frac{1}{\sqrt{n}}a^T w - 2 \sqrt{c(1-c)})^2 + \lambda \|w\|^2 
\end{align}

Note that $(g^T w + \sqrt{n} \sigma \|w\|)_{\ge 0}^2 \le  n\sigma^2 \|w\|^2 \ll \lambda \|w\|^2$ if  $\lambda \gg n\sigma^2$. Thus, the term $(g^T w + \sqrt{n} \sigma \|w\|)_{\ge 0}^2$ is dominated by the regularization term, which is why we can drop it in this regime. After doing so, we get the following objective:
\begin{align*}
    \min_w \frac{n}{2} (\Tilde{\mu_1}^T w - (1-2c))^2 + \frac{n}{2} (\Tilde{\mu_2}^T w + (1-2c))^2 + n (\frac{1}{\sqrt{n}}a^T w - 2 \sqrt{c(1-c)})^2 + \lambda \|w\|^2 
\end{align*}
We claim that $w$ can be written in the form $w = \alpha g + \beta a + \gamma_1\Tilde{\mu_1} + \gamma_2\Tilde{\mu_2}$. Indeed, $w$ belongs in the span of $g, a, \mu_1$ and $\mu_2$ because all terms of (\ref{eq: ao_detailed_ridge}) but the last one depend only on the projections of $w$ onto these directions and the last term is does not increase if all other directions are dropped from $w$. Note also that $\gamma_1 = -\gamma_2 =: \gamma$ because (\ref{eq: ao_detailed_ridge}) is invariant under the transformation $(\gamma_1, \gamma_2, \mu_1, \mu_2) \to (-\gamma_2, \gamma_1, -\mu_2, \mu_1)$. Hence, $w = \alpha g + \beta a + \gamma(\Tilde{\mu_1}-\Tilde{\mu_2})$.

Taking derivative of (\ref{eq: ao_detailed_ridge}) w.r.t $\beta$:
\begin{align*}
    & 2\sqrt{n} d \sigma^2 ( \frac{d \sigma^2}{\sqrt{n}} \beta  - 2 \sqrt{c(1-c)}) + 2 \lambda d \sigma^2 \beta= 0 \\
    &  \sqrt{n} ( \frac{d \sigma^2}{\sqrt{n}} \beta - 2 \sqrt{c(1-c)}) +  \lambda \beta = 0 \\ 
     &  (d \sigma^2 + \lambda) \beta - 2\sqrt{nc(1-c)} = 0 \\
\end{align*}
For $\gamma$:
\begin{align*}
    & n ( d (1-r + \frac{2\sigma^2}{n}) \gamma  - 1 + 2c) + 2 \lambda \gamma = 0\\ 
    &(dn (1-r + \frac{2\sigma^2}{n}) + 2\lambda) \gamma - n(1 - 2c) = 0
\end{align*}
We thus obtain:
\begin{align*}
& \beta = \frac{2\sqrt{nc(1-c)}}{d \sigma^2 + \lambda}\\
    & \gamma = \frac{1-2c}{d(1-r + \frac{2\sigma^2}{n}) + 2\frac{\lambda}{n}}
\end{align*}

To calculate the generalization error, note that $\|w\|^2 = d (\alpha^2  \sigma^2 + \beta^2 \sigma^2 + 2\gamma^2 (1-r + \frac{2\sigma^2}{n}))$ and $\Tilde{\mu_1}^T w = d \gamma (1-r + \frac{2\sigma^2}{n}))$. Since we deemed the effect of the first term in (\ref{eq: ao_detailed_ridge}) negligible, we can set $\alpha = 0$, as this parameter does not play any role. The error then equals to the following:
\begin{align*}
    E(w) = Q (\frac{d(1-r) \gamma }{\sigma \sqrt{2 d (1-r+\frac{2\sigma^2}{n}) \gamma^2 + d \sigma^2 \beta ^2}})  
\end{align*}

\subsection{Derivations behind Remark \ref{rem: l2Qfunc}}

    Assume $\sigma \ll \sqrt{n}$ and note that $$\frac{d(1-r) \gamma }{\sigma \sqrt{2 d (1-r+\frac{2\sigma^2}{n})\gamma^2 + d \sigma^2 \beta ^2}} = \frac{d(1-r)  }{\sigma \sqrt{2 d (1-r+\frac{2\sigma^2}{n}) + d \sigma^2 (\frac{\beta}{\gamma}) ^2}} \approx \frac{d(1-r)  }{\sigma \sqrt{2 d (1-r) + d \sigma^2 (\frac{\beta}{\gamma}) ^2}}$$

Plugging in the expressions for $\beta$ and $\gamma$ from the previous remark, we have $$ \displaystyle \frac{\beta}{\gamma} \xrightarrow[\lambda \to +\infty ]{} \frac{\sqrt{c(1-c)}(1-2c)}{\sqrt{n}}$$

Denoting $C = \sqrt{c(1-c)}(1-2c)$, we derive

$$\frac{d(1-r)  }{\sigma \sqrt{2 d (1-r) + d \sigma^2 (\frac{\beta}{\gamma}) ^2}} \xrightarrow[\lambda \to +\infty ]{} \frac{d(1-r)  }{\sigma \sqrt{2 d (1-r) +  \frac{d \sigma^2C^2}{n}}}$$

Since $\sigma \ll \sqrt{n}$, the last term in the denominator can be neglected and the expression can be approximated as 
$$\frac{d(1-r)  }{\sigma \sqrt{2 d (1-r) }} = \frac{\sqrt{d(1-r)} }{\sigma \sqrt{2}}$$

\subsection{Derivations behind Remark \ref{rem:l_1_sparse}}
In this section we present the optimal values of $\gamma$ and $\beta$ which are used in the statement of Remark \ref{rem:l_1_sparse} and show that indeed the optimal $w_i$ attains the value $0$ except when $|t_i| :=|(\frac{n}{2} \gamma(\Tilde{\mu_1} - \Tilde{\mu_2}) + \sqrt{n} \beta a)_i| \geq \lambda$. Let us consider the following optimization which comes from the equation (\ref{eq: ao_detailed}):
\begin{align*}
    \min_w (g^T w + \sqrt{n} \sigma \|w\|)_{\ge 0}^2 + \frac{n}{2} (\Tilde{\mu_1}^T w - (1-2c))^2 + \frac{n}{2} (\Tilde{\mu_2}^T w + (1-2c))^2 + n (\frac{1}{\sqrt{n}}a^T w - 2 \sqrt{c(1-c)})^2 + \lambda \|w\|_1
\end{align*}
For large enough $\lambda$, in a similar argument to the ridge regression case, we can neglect the first term because it is dominated by the regularization term, which yields:
\begin{align*}
    \min_w  \frac{n}{2} (\Tilde{\mu_1}^T w - (1-2c))^2 + \frac{n}{2} (\Tilde{\mu_2}^T w + (1-2c))^2 + n (\frac{1}{\sqrt{n}}a^T w - 2 \sqrt{c(1-c)})^2 + \lambda \|w\|_1
\end{align*}
Now we introduce $\gamma_1, \gamma_2, \beta$ through taking the Fenchel Dual of the square terms:
\begin{align*}
    \min_w \max_{\gamma_1, \gamma_2, \beta} \frac{n}{2} \gamma_1(\Tilde{\mu_1}^T w - (1-2c)) -\frac{n\gamma_1^2}{8} + \frac{n}{2} \gamma_2(\Tilde{\mu_2}^T w + (1-2c)) -\frac{n\gamma_2^2}{8} + n \beta (\frac{1}{\sqrt{n}}a^T w - 2 \sqrt{c(1-c)}) -\frac{n\beta^2}{4} + \lambda \|w\|_1 
\end{align*}
Rearranging the order of optimization to find the optimal w:
\begin{align*}
    &\max_{\gamma_1, \gamma_2, \beta} -\frac{n}{2} (1-2c) \gamma_1 -\frac{n\gamma_1^2}{8} +\frac{n}{2} (1-2c) \gamma_2 -\frac{n\gamma_1^2}{8} - 2n  \sqrt{c(1-c)} \gamma -\frac{n\beta^2}{4} + \\
    &+ \sum_{i=1}^d \min_{w_i} ( \frac{n}{2} \gamma_1 \Tilde{\mu_1} + \frac{n}{2} \gamma_2 \Tilde{\mu_2} + \sqrt{n} \beta a)_i w_i + \lambda |w_i|
\end{align*}
The optimization over $w$ reduces to:
\begin{align*}
    \min_{w_i} c_i w_i + \lambda |w_i| = \begin{cases}
        0 & |c_i| < \lambda \\
        0 & |c_i| = \lambda \\
        -\infty & |c_i| > \lambda
    \end{cases}
\end{align*}
Therefore the solution would be:
\begin{align*}
    w_{opt,i}  = \begin{cases}
        0 & |c_i| < \lambda \\
        [0, \infty) sign(-c_i) & |c_i| = \lambda \\
        \infty sign(-c_i) & |c_i| > \lambda
    \end{cases}
\end{align*}
Note that plugging in $w_{opt,i}$ results in adding the following constraint: $$|(\frac{n}{2} \gamma_1 \Tilde{\mu_1} + \frac{n}{2} \gamma_2 \Tilde{\mu_2} + \sqrt{n} \beta a)_i|\leq \lambda$$ 
Or overall, $$\max_i |(\frac{n}{2} \gamma_1 \Tilde{\mu_1} + \frac{n}{2} \gamma_2 \Tilde{\mu_2} + \sqrt{n} \beta a)_i|\leq \lambda$$ 
Which is equivalent to: 
$$\|\frac{n}{2} \gamma_1 \Tilde{\mu_1} + \frac{n}{2} \gamma_2 \Tilde{\mu_2} + \sqrt{n} \beta a\|_{\infty}\leq \lambda$$

Denote 
$$\Psi := \frac{n}{2} \gamma_1 \Tilde{\mu_1} + \frac{n}{2} \gamma_2 \Tilde{\mu_2} + \sqrt{n} \beta a$$ 

Using the preceding discussion, the optimization can be written as:
\begin{align*}
    \max_{\substack{\gamma_1, \gamma_2, \beta \\ \|\Psi\|_{\infty}\leq \lambda} } -\frac{n}{2} (1-2c) \gamma_1 -\frac{n\gamma_1^2}{8} +\frac{n}{2} (1-2c) \gamma_2 -\frac{n\gamma_1^2}{8} - 2n  \sqrt{c(1-c)} \beta-\frac{n\beta^2}{4}
\end{align*}
Furthermore, using the assumptions on $\mu_1, \mu_2$, one can observe: 
$$(\frac{n}{2} \gamma_1 \Tilde{\mu_1} + \frac{n}{2} \gamma_2 \Tilde{\mu_2} + \sqrt{n} \beta a)_i \stackrel{d}{=} \sqrt{\frac{n^2}{4}(1+\frac{2\sigma^2}{n}) (\gamma_1^2+\gamma_2^2)+\frac{n^2}{2}r \gamma_1 \gamma_2 + n \sigma^2 \beta^2} G_i$$
where $G_i \sim \mathcal{N}(0,1)$ and $\stackrel{d}{=}$ stands for equality in distribution. 

Denote $q := \sqrt{\frac{n^2}{4}(1+\frac{2\sigma^2}{n}) (\gamma_1^2+\gamma_2^2)+\frac{n^2}{2}r \gamma_1 \gamma_2 + n \sigma^2 \beta^2}$.
Then the equivalent optimization would be:
\begin{align*}
    \max_{\substack{\gamma_1, \gamma_2, \beta\\ q \mathbb{E}\|G\|_{\infty}\leq \lambda} } -\frac{n}{2} (1-2c) \gamma_1 -\frac{n\gamma_1^2}{8} +\frac{n}{2} (1-2c) \gamma_2 -\frac{n\gamma_1^2}{8} - 2n  \sqrt{c(1-c)} \beta-\frac{n\beta^2}{4}
\end{align*}
We know that $\|X\|_{\infty}$ concentrates to  $\mathbb{E}\|X\|_{\infty} = \sqrt{2 \log(2d)}$ for i.i.d. standard Gaussians $X = (X_1, X_2, ..., X_d)$ (cf. Gumbel distribution for Gaussian random variables). Then, using a Lagrange multiplier:
\begin{align*}
    & \max_{\gamma_1, \gamma_2, \beta} \min_{\eta \geq 0}  -\frac{n}{2} (1-2c) \gamma_1 -\frac{n\gamma_1^2}{8} +\frac{n}{2} (1-2c) \gamma_2 -\frac{n\gamma_1^2}{8} - 2n  \sqrt{c(1-c)} \beta-\frac{n\beta^2}{4} + \\ & + \eta \left(\lambda^2 - 2 \left[ \frac{n^2}{4}(1+\frac{2\sigma^2}{n}) (\gamma_1^2+\gamma_2^2)+\frac{n^2}{2}r \gamma_1 \gamma_2 + n \sigma^2 \beta^2  \right] \log(2d) \right)
\end{align*}
Convexity in $\eta$ and concavity in all other variables allow us to swap min and max:
\begin{align*}
    &\min_{\eta \geq 0} \lambda^2 \eta + + \max_{\gamma_1, \gamma_2, \beta}\\
    &  - \begin{pmatrix}
        \gamma_1 & \gamma_2 & \beta& 1
    \end{pmatrix} \begin{pmatrix}
        \frac{n}{8} + \frac{n^2}{2} \eta (1+\frac{2\sigma^2}{n}) \log(2d) & \frac{n^2}{2}r \eta \log(2d) & 0  & \frac{n}{4} (1-2c) \\
        \frac{n^2}{2}r \eta \log(2d) & \frac{n}{8} + \frac{n^2}{2} \eta (1+\frac{2\sigma^2}{n}) \log(2d) & 0 & -\frac{n}{4} (1-2c) \\
        0 & 0 & \frac{n}{4} + 2n \sigma^2 \eta \log(2d) & n  \sqrt{c(1-c)} \\
        \frac{n}{4} (1-2c) & -\frac{n}{4} (1-2c)& n  \sqrt{c(1-c)} & 0
    \end{pmatrix} \begin{pmatrix}
        \gamma_1 \\
        \gamma_2 \\
        \beta\\
        1
    \end{pmatrix}
\end{align*}
Note that
\begin{align*}
    \frac{1}{a^2-b^2}
    \begin{pmatrix}
        x & -x
    \end{pmatrix}
    \begin{pmatrix}
        a & -b \\
        -b & a
    \end{pmatrix}
    \begin{pmatrix}
        x \\ -x
    \end{pmatrix} = \frac{x^2}{a^2-b^2} \begin{pmatrix}
        a+b & -b-a
    \end{pmatrix}\begin{pmatrix}
        1 \\ -1
    \end{pmatrix} = \frac{2a+2b}{a^2-b^2}x^2 = \frac{2x^2}{a-b}
\end{align*}
To do the quadratic optimization, we calculate the Schur Complement:
\begin{align*}
    & \min_{\eta \geq 0} \lambda^2 \eta + \begin{pmatrix}
         \frac{n}{4} (1-2c) & -\frac{n}{4} (1-2c) & n  \sqrt{c(1-c)}
     \end{pmatrix}  \cdot \\ &
     \cdot  \begin{pNiceMatrix}[margin]
        \frac{n}{8} + \frac{n^2}{2} \eta (1+\frac{2\sigma^2}{n}) \log(2d) & \frac{n^2}{2}r \eta \log(2d) & 0 \\
        \frac{n^2}{2}r \eta \log(2d) & \frac{n}{8} + \frac{n^2}{2} \eta (1+\frac{2\sigma^2}{n}) \log(2d) & 0 \\
         0 & 0 & (\frac{n}{4} + 2n \sigma^2 \eta \log(2d))^{-1} \\
         \CodeAfter
         \SubMatrix({1-1}{2-2})^{-1}
    \end{pNiceMatrix}
    \begin{pmatrix}
         \frac{n}{4} (1-2c) \\ -\frac{n}{4} (1-2c) \\ n  \sqrt{c(1-c)}
     \end{pmatrix}    
\end{align*}
We note that indeed $\gamma_1 = - \gamma_2 =: \gamma$ and subsequently,
\begin{align*}
    \min_{\eta \geq 0} \lambda^2 \eta + \frac{n^2}{8} (1-2c)^2 \frac{1}{\frac{n}{8} + \frac{n^2}{2} \eta (1-r+\frac{2\sigma^2}{n}) \log(2d)} + \frac{n^2 c (1-c)}{\frac{n}{4} + 2n \sigma^2 \eta \log(2d)}
\end{align*}
Consequently, to find $\gamma, \beta$ it suffices to solve the following 1-variable optimization
\begin{align*}
    \min_{\eta \geq 0} \lambda^2 \eta + \frac{n (1-2c)^2}{1 + 4n \eta (1-r+\frac{2\sigma^2}{n}) \log(2d)} + \frac{4n c (1-c)}{1 + 8 \sigma^2 \eta \log(2d)}
\end{align*}
Having found $\eta$ from the above optimization, we can plug in the value of $\eta$ to find $\gamma, \beta$:
\begin{align*}
    \beta= -  \frac{4\sqrt{c(1-c)}}{1 + 8 \sigma^2 \eta \log(2d)}, \quad \gamma = - \frac{2 (1-2c)}{1 + 4n \eta (1-r+\frac{2\sigma^2}{n}) \log(2d)}
\end{align*}
By going through this machinery, we reduced a complicated optimization over 3 variables, to a rather simple optimization over 1 variable as $\eta$ is the solution to a degree 4 equation.

\subsection{Derivations behind  Remark \ref{rem:l_inf_compr}}
Similar to the previous section, we present an approach through which, $\gamma, \beta$ can be calculated efficiently. By taking $\lambda$ to be large enough, we can drop the first term which in turn yields:
\begin{align*}
    \min_w  \frac{n}{2} (\Tilde{\mu_1}^T w - (1-2c))^2 + \frac{n}{2} (\Tilde{\mu_2}^T w + (1-2c))^2 + n (\frac{1}{\sqrt{n}}a^T w - 2 \sqrt{c(1-c)})^2 + \lambda \|w\|_{\infty}
\end{align*}
Adding a new variable $\delta$ for $\lambda \|w\|_{\infty}$ we obtain:
\begin{align*}
    \min_{\|w\|_{\infty} \leq \frac{\delta}{\lambda}, \delta \geq 0} \frac{n}{2} (\Tilde{\mu_1}^T w - (1-2c))^2 + \frac{n}{2} (\Tilde{\mu_2}^T w + (1-2c))^2 + n (\frac{1}{\sqrt{n}}a^T w - 2 \sqrt{c(1-c)})^2 + \delta 
\end{align*}
Introducing $\gamma_1, \gamma_2$ through Fenchel Dual and changing the order of optimization we derive the following objective:
\begin{align*}
    &\min_{\delta \geq 0 } \delta + \max_{\gamma_1, \gamma_2, \beta} -\frac{n}{2} (1-2c) \gamma_1 -\frac{\gamma_1^2}{4} +\frac{n}{2} (1-2c) \gamma_2 -\frac{\gamma_2^2}{4} - 2n  \sqrt{c(1-c)} \beta-\frac{\beta^2}{4} + \\
    &+ \sum_{i=1}^d \min_{|w_i| \leq \frac{\delta}{\lambda}} ( \frac{n}{2} \gamma_1 \Tilde{\mu_1} + \frac{n}{2} \gamma_2 \Tilde{\mu_2} + \sqrt{n} \beta a)_i w_i
\end{align*}
It is straightforward to see that $w_{opt, i} = - \frac{\delta}{\lambda} sign(( \frac{n}{2} \gamma_1 \Tilde{\mu_1} + \frac{n}{2} \gamma_2 \Tilde{\mu_2} + \sqrt{n} \beta a)_i)$. So, the optimization would be
\begin{align*}
    &\min_{\delta \geq 0 } \delta + \max_{\gamma_1, \gamma_2, \beta} -\frac{n}{2} (1-2c) \gamma_1 -\frac{\gamma_1^2}{4} +\frac{n}{2} (1-2c) \gamma_2 -\frac{\gamma_1^2}{4} - 2n  \sqrt{c(1-c)} \beta -\frac{\beta^2}{4} + \\
    &- \frac{\delta}{\lambda} \| \frac{n}{2} \gamma_1 \Tilde{\mu_1} + \frac{n}{2} \gamma_2 \Tilde{\mu_2} + \sqrt{n} \beta a\|_1
\end{align*}
Let $\Psi:= \frac{n}{2} \gamma_1 \Tilde{\mu_1} + \frac{n}{2} \gamma_2 \Tilde{\mu_2} + \sqrt{n} \beta a$. Optimizing over $\delta$ yields:
\begin{align*}
    \max_{\substack{\gamma_1, \gamma_2, \beta \\ \|\Psi\|_{1}\leq \lambda} } -\frac{n}{2} (1-2c) \gamma_1 -\frac{\gamma_1^2}{4} +\frac{n}{2} (1-2c) \gamma_2 -\frac{\gamma_1^2}{4} - 2n  \sqrt{c(1-c)} \beta -\frac{\beta^2}{4}
\end{align*}
In a manner similar to the previous section we note: $$(\frac{n}{2} \gamma_1 \Tilde{\mu_1} + \frac{n}{2} \gamma_2 \Tilde{\mu_2} + \sqrt{n} \beta a)_i \stackrel{d}{=} \sqrt{\frac{n^2}{4}(1+\frac{2\sigma^2}{n}) (\gamma_1^2+\gamma_2^2)+\frac{n^2}{2}r \gamma_1 \gamma_2 + n \sigma^2 \beta^2 } G_i$$ with $G_i \sim \mathcal{N}(0,1)$. Denoting, $q:= \sqrt{\frac{n^2}{4}(1+\frac{2\sigma^2}{n}) (\gamma_1^2+\gamma_2^2)+\frac{n^2}{2}r \gamma_1 \gamma_2 + n \sigma^2 \beta^2 }$ This result entails the following objective:
\begin{align*}
    \max_{\substack{\gamma_1, \gamma_2, \beta \\  q\mathbb{E}\|G\|_{1}\leq \lambda} } -\frac{n}{2} (1-2c) \gamma_1 -\frac{\gamma_1^2}{4} +\frac{n}{2} (1-2c) \gamma_2 -\frac{\gamma_1^2}{4} - 2n  \sqrt{c(1-c)} \beta -\frac{\beta^2}{4}
\end{align*}
We leverage the fact that for iid standard Gaussians $X = (X_1, X_2, ..., X_d)$, $\mathbb{E} \|X\|_{1} \sim d\sqrt{\frac{2}{\pi}}$. To write the optimization in a more tractable format, we use a Lagrange multiplier to bring in the constraint:
\begin{align*}
    & \max_{\gamma_1, \gamma_2, \beta} \min_{\eta \geq 0}  -\frac{n}{2} (1-2c) \gamma_1 -\frac{\gamma_1^2}{4} +\frac{n}{2} (1-2c) \gamma_2 -\frac{\gamma_1^2}{4} - 2n  \sqrt{c(1-c)} \beta -\frac{\beta^2}{4} + \\ & + \eta \left(\lambda^2 - 2 \left[ \frac{n^2}{4}(1+\frac{2\sigma^2}{n}) (\gamma_1^2+\gamma_2^2)+\frac{n^2}{2}r \gamma_1 \gamma_2 + n \sigma^2 \beta^2  \right] \frac{d^2}{\pi} \right)
\end{align*}
It can be seen that we arrive at an objective which is essentially the same as the one in the previous section, but with $\log(2d)$ being replaced by $\frac{d^2}{\pi}$. Therefore we can use the results from before and derive the final optimization for $\eta$:
\begin{align*}
    \min_{\eta \geq 0} \lambda^2 \eta + \frac{n^2 (1-2c)^2}{2 + 4n^2 \eta (1-r+\frac{2\sigma^2}{n}) \frac{d^2}{\pi}} + \frac{4n^2 c (1-c)}{1 + 8n \sigma^2 \eta \frac{d^2}{\pi}}
\end{align*}
We note that $\gamma_1 = -\gamma_2 =: \gamma$. Having found the $\eta$, we therefore obtain:
\begin{align*}
    \beta= -  \frac{4\sqrt{c(1-c)}}{1 + 8 \sigma^2 \eta \frac{d^2}{\pi}}, \quad \gamma = - \frac{2 (1-2c)}{1 + 4n \eta (1-r+\frac{2\sigma^2}{n}) \frac{d^2}{\pi}}
\end{align*}

\section{SOLVING THE SCALARIZED AO NUMERICALLY}

Having simplified a high-dimensional optimization problem to a low-dimensional, i.e. one having 2-3 variables, the task of solving the scalarized optimization remains. Since the objective from the conclusion of Theorem \ref{thm: ridge} is a convex minimization problem, it could be handled directly using CVX. However, this is not the case for Theorems \ref{thm: AO_l1} and \ref{thm: AO_linf}, where one must deal with a $\min\max$. Since there are no packages available for solving  $\min\max$ problems numerically, we needed to adapt different approaches. For Theorem  \ref{thm: AO_l1}, we conducted a grid search over $\tau$ and for each $\tau$ we performed maximization over $\beta$ and $\gamma$ using MATLAB fmincon. Alternatively, doing a grid search over all variables is also a feasible approach, which may not work effectively when the number of variables and the number of points in the grid increase. Nevertheless, this is what we used for simulating Theorem \ref{thm: AO_linf}, which turned out to work well in practice for $3$ variables and a grid with $8000000$ points.

\end{document}